\documentclass[lettersize,journal]{IEEEtran}
\usepackage{amsmath,amsfonts}
\usepackage{algorithmic}
\usepackage{algorithm}
\usepackage{array}
\usepackage[caption=false,font=normalsize,labelfont=sf,textfont=sf]{subfig}
\usepackage{textcomp}
\usepackage{stfloats}
\usepackage{adjustbox}
\usepackage{url}
\usepackage{verbatim}
\usepackage{graphicx}
\usepackage{cite}
\usepackage{multirow}
\usepackage{amsmath}
\usepackage{amssymb}
\usepackage{amsthm}
\usepackage{booktabs}
\usepackage{color,xcolor}
\usepackage{algorithm}
\usepackage{algorithmic}
\usepackage{threeparttable}

\usepackage{newfloat}
\usepackage{listings}
\DeclareCaptionStyle{ruled}{labelfont=normalfont,labelsep=colon,strut=off} 
\floatstyle{ruled}
\newfloat{listing}{tb}{lst}{}
\floatname{listing}{Listing}
\usepackage[capitalize]{cleveref}

\begin{document}

\title{RARE: Robust Masked Graph Autoencoder}

\author{Wenxuan Tu,
        Qing Liao,
        Sihang Zhou,
        Xin Peng,
        Chuan Ma,
        Zhe Liu,~\IEEEmembership{Senior Member, IEEE},\\
        Xinwang Liu$^{\ast}$,~\IEEEmembership{Senior Member, IEEE},
        and Zhiping Cai
        
\thanks{W. Tu and X. Liu are with the College of Computer, National University of Defense Technology, Changsha 410073, China (e-mail: wenxuantu@163.com, xinwangliu@nudt.edu.cn).} 
\thanks{$^{\ast}$ Corresponding author.}
}

\markboth{}%
{Shell \MakeLowercase{\textit{et al.}}: A Sample Article Using IEEEtran.cls for IEEE Journals}

\maketitle

\begin{abstract}
Masked graph autoencoder (MGAE) has emerged as a promising self-supervised graph pre-training (SGP) paradigm due to its simplicity and effectiveness. However, existing efforts perform the mask-then-reconstruct operation in the raw data space as is done in computer vision (CV) and natural language processing (NLP) areas, while neglecting the important non-Euclidean property of graph data. As a result, the highly unstable local connection structures largely increase the uncertainty in inferring masked data and decrease the reliability of the exploited self-supervision signals, leading to inferior representations for downstream evaluations. To address this issue, we propose a novel SGP method termed \textbf{\underline{R}}obust m\textbf{\underline{A}}sked g\textbf{\underline{R}}aph auto\textbf{\underline{E}}ncoder (RARE) to improve the certainty in inferring masked data and the reliability of the self-supervision mechanism by further masking and reconstructing node samples in the high-order latent feature space. Through both theoretical and empirical analyses, we have discovered that performing a joint mask-then-reconstruct strategy in both latent feature and raw data spaces could yield improved stability and performance. To this end, we elaborately design a masked latent feature completion scheme, which predicts latent features of masked nodes under the guidance of high-order sample correlations that are hard to be observed from the raw data perspective. Specifically, we first adopt a latent feature predictor to predict the masked latent features from the visible ones. Next, we encode the raw data of masked samples with a momentum graph encoder and subsequently employ the resulting representations to improve predicted results through latent feature matching. Extensive experiments on seventeen datasets have demonstrated the effectiveness and robustness of RARE against state-of-the-art (SOTA) competitors across three downstream tasks. 
\end{abstract}


\section{Introduction}
\IEEEPARstart{M}{asked} autoencoders (MAEs) have emerged as the dominant technique for self-supervised vision and language pre-training. The objective of MAEs is to learn generalized sample representations from massive unlabeled data by recovering partially masked content (\textit{e.g.,} image patches or word embeddings) from observations. Due to their simplicity and powerful local structure modeling capabilities, advanced efforts in this field \cite{2022MAE,2022data2vec,2022CAE} have garnered significant interest among researchers. These methods have demonstrated impressive performance across a wide range of real-world applications, including medical image analysis \cite{2022M3AE}, natural language understanding \cite{2022TACO}, and 3D object detection \cite{2023PiMAE}.

\begin{figure}[!t]
\centering
\includegraphics[width=3.4in]{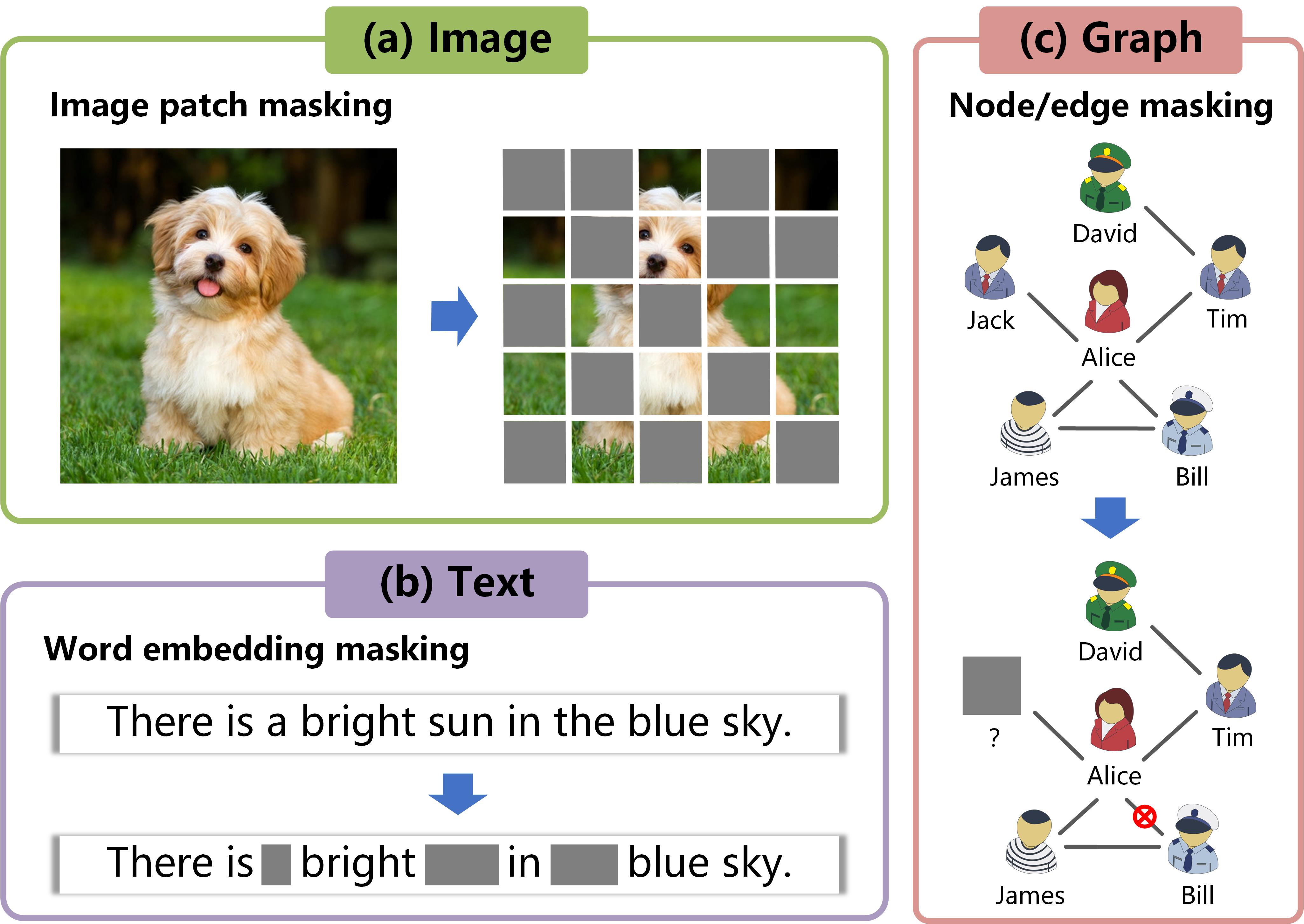}
\caption{Information masking on different types of data. For instance, masking image patches on images (a) or word embeddings on texts (b) would not alter the underlying semantics of the original data to some extent. In other words, even after a portion of image patches or word embeddings are masked, people can still recognize the object visually or understand the language content by inferring the invisible content based on observed contexts. However, masking nodes or edges on non-euclidean graphs (c) may unexpectedly increase the uncertainty when inferring masked data, since the nearby structure of a node or an edge is less stable and has lower certainty than that of an image or a text.}
\label{1}
\end{figure}

Recent studies have shown that applying MAEs to facilitate graph-based learning has become a topic of increasing interest. The success of masked graph autoencoders (MGAEs) lies in the mask-then-reconstruct operation. In this setting, a portion of visible nodes or edges are randomly masked and then adopted as self-supervision signals to guide the model learning, so as to allow the network to explore the underlying structural information for downstream evaluations \cite{2022MGAE, 2022MaskGAE, 2022GraphMAE, 2022HGMAE, 2022MGAP, 2022BatmanNet, 2023SeeGera}. Despite their promising performance on various graph-oriented tasks \cite{2023EMSTGAE,2022MGMAE,2022NAS}, existing MGAEs overlook the inherent distinction between images (or texts) and graphs, \textit{i.e.,} images (or texts) are Euclidean while graphs are non-Euclidean. In other words, the nearby structure of an image patch or a sub-sentence has higher semantic certainty and is more stable than that of a sub-graph. Therefore, the quality of the self-supervision guidance provided by a masked image patch or a masked word embedding would be much higher than that provided by a masked node (or edge). Specifically, as shown in Fig. \ref{1}(a), since the relative spatial distribution of organs on a dog is quite certain, people can easily imagine the masked image patches based on the observed content within an incomplete dog photo. Similarly, in Fig. \ref{1}(b), the strong context correlation among words helps us involuntarily fill an incomplete sentence with meaningful words. Comparatively, in a social network where nodes are entities and edges are interactions, the neighborhood structure of an entity within a graph varies a lot, as shown in Fig. \ref{1}(c). When a node or an edge is entirely masked, it is hard to identify the removed entity or ascertain whether two entities keep in contact directly, since it is common for two unconnected entities to have conjoint neighbors or valuable higher-order relationships that are hard to be observed from the raw data perspective in a real-world social graph. Consequently, although in most cases, the mask-then-reconstruct principle is effective for learning valuable node representations, directly recovering the masked nodes and edges driven by the low-level raw data would put the corresponding model at risk of being confused by the local structural ambiguity. Based on these observations, we argue that the non-Euclidean property of graph data could to some extent trigger uncertainty in inferring masked data and may negatively affect the reliability of the self-supervision mechanism. In this circumstance, the robustness of the model may be compromised when applying a masked autoencoder to process graphs straightforwardly. 

To address the above issue, we propose a novel method termed \textbf{\underline{R}}obust m\textbf{\underline{A}}sked g\textbf{\underline{R}}aph auto\textbf{\underline{E}}ncoder (RARE) for self-supervised graph pre-training. The main idea of RARE is to integrate implicit and explicit self-supervision mechanisms for masked content recovery by performing a joint mask-then-reconstruct strategy in both latent feature and raw data spaces. The effectiveness of our method lies in the fact that unlike the model optimization driven by the low-level raw data only \cite{2022MaskGAE,2022GraphMAE}, the self-supervision mechanism of RARE could be further enhanced by incorporating more informative high-order sample correlations that are hard to be observed from the raw data perspective \cite{2019Higher-order,2023MAGC,2023HAE-GNN}. To this end, we design a masked latent feature completion scheme that includes two steps. Specifically, we first adopt a latent feature predictor to assist the graph encoder in extracting more compressed features by predicting the latent features of masked samples based on observations. To further enhance the integrity and accuracy of predicted representations, we encode the raw data of masked samples using a momentum graph encoder and leverage the resultant representations to guide the latent feature prediction through information matching. With such persistent and informative signals as self-supervision guidance, each masked sample in the latent space is encouraged to explore reliable information from available features. As the implicit self-supervision signals become more accurate and the masked content becomes more reliable, the model is enforced to promote greater information encoding capability, thereby generating more robust and generalized node representations for downstream evaluations. The main contributions of this work are summarized as below: 

\begin{itemize}
\item We propose a novel SGP framework termed RARE to enhance the robustness of masked graph autoencoders. It not only eases the instability of the self-supervision mechanism driven by the non-Euclidean raw graph data, but also achieves a good accuracy-efficiency trade-off. 

\item We incorporate a simple but effective masked latent feature completion scheme into a masked graph autoencoder. This design can enhance the certainty in inferring masked data and the reliability of the self-supervision mechanism by exploiting more informative high-order sample correlations to drive the model training. 

\item Extensive experiments on seventeen datasets across three downstream tasks demonstrate the effectiveness and robustness of RARE over competitors. Moreover, a series of elaborate ablation studies also verify that RARE can indeed unleash the full potential of MAEs to provide a comprehensive understanding of graphs. 
\end{itemize}

The remainder of this paper is organized as follows. In Section II, we review related work in areas of self-supervised graph pre-training, masked graph autoencoders, and self-distillation on graphs. Section III presents defined notations, the proposed network designs, loss functions, and discussions. In Section IV, we conduct experiments and analyze the results. Section V draws a final conclusion.

\section{Related Work}
\label{sec:rela}
\subsection{Self-supervised Graph Pre-training}
Self-supervised graph pre-training (SGP), whose goal is to learn representations from supervised signals derived from the graph data itself, has made significant progress recently. With the powerful learning capability of graph neural networks (GNNs) \cite{2022Survey,2017GCN,2018GAT,2018GIN}, advanced studies in this field have recently achieved great success in anomaly detection \cite{2021HO-GAT}, feature selection \cite{2023MGAGR}, multi-view clustering \cite{2019GSF,2022SMVC}, etc. One of the most representative self-supervised learning paradigms is contrastive SGP, where discriminative features are learned by pulling together the representations of semantically similar samples while pushing away the ones of unrelated samples \cite{2019DGI,2020GraphCL,2020GCC,2020MVGRL,2020GRACE,2020Infograph,2021CCA--SSG,2023AutoProNE}. However, the impressive performance of these methods heavily relies on careful data augmentations, large amounts of negative samples, or relatively complicated optimization strategies, which usually causes time- and resource-consuming issues. Comparatively, generative SGP methods \cite{2016VGAE,2020SDCN,2021DFCN,2022ITR,2022GraphMAE,2023TTER} could naturally avoid the low-efficiency problem, as their optimization target is to reconstruct the input (or masked) information directly. In particular, masked graph autoencoders (MGAEs) \cite{2022GraphMAE}, which aim to predict the masked content from visible one, have significantly advanced classical graph autoencoders and shown their potential to achieve better performance against contrastive learning-based competitors. 

\subsection{Masked Graph Autoencoders}
Masked signal modeling (MSM), which models masked signals locally to facilitate the extraction of significant features, has recently gained popularity in self-supervised vision and language applications, such as natural language understanding \cite{2022TACO} and medical image analysis \cite{2022M3AE}. Inspired by the successes of existing masked autoencoders (MAEs) \cite{2019BERT,2022MAE,2022CAE,2022data2vec}, researchers pose a natural question regarding the potential of utilizing MAEs to handle large amounts of unlabelled graph data. To this end, MGAE \cite{2022MGAE} first applies an undirected edge-masking strategy to the original graph structure, and then utilizes a tailored cross-correlation decoder to predict the masked edges via a standard graph-based loss function. Similarly, MaskGAE \cite{2022MaskGAE} incorporates random corruption into the graph structure from both edge-wise level and path-wise level, and then utilizes edge-reconstruction and node-regression loss functions to match the predicted information with the original data. Another important research line in this field is node-masking-based MGAEs \cite{2022GMAE,2022GraphMAE,2022HGMAE}. For example, GMAE \cite{2022GMAE} utilizes a graph transformer-based backbone \cite{2019GTN} to pre-train a masked autoencoder, and applies a cross-entropy loss to compare the reconstructed attributes with their ground truths. Similarly, GraphMAE \cite{2022GraphMAE} reconsiders the reconstruction loss functions of previous graph autoencoders and proposes an improved scaled cosine error to boost the masked attribute recovery. More recently, some studies first randomly mask a portion of node attributes and connections of given graphs simultaneously, and then learn to predict the removed content from available information via two specific reconstruction objectives \cite{2022BatmanNet,2023SeeGera}. However, the aforementioned methods usually learn representations by directly minimizing a reconstruction loss in the raw data space, which may mislead the model into a local structural ambiguity situation caused by the non-Euclidean property of graphs. In contrast, RARE can effectively enhance the certainty in inferring masked data and the reliability of the self-supervision mechanism by reconciling the reconstructions of masked raw attributes and latent features, unleashing the potential of MAEs for graph analysis while mitigating the aforementioned negative effects. 

\subsection{Self-distillation on Graphs}
Self-distillation has emerged as a powerful technique for self-supervised learning, as evidenced by its successful applications in various domains\cite{2017MeanTeacher,2018DeepClustering,2020BYOL,2021BGRL,2022SUBLIME}. This technique involves using the outputs of a target network as pseudo labels to guide the representation learning of an online network, which encourages the feature extractor to learn more generalized features. Self-distillation has also been widely developed and employed in multiple graph machine learning tasks, including graph structure learning \cite{2022SUBLIME} and augmentation-free node clustering \cite{2022AFGRL}. While previous studies have focused on complete graphs, it is crucial to explore the effectiveness of self-distillation in boosting MGAEs in scenarios where the graph data is incomplete. This motivates us to investigate the feasibility of enhancing the robustness of MGAEs with the aid of self-distillation learning, as well as uncovering the key factors that contribute to the success of RARE.

\section{Method}
\begin{table}[!t]
    \centering
    \scriptsize
    \caption{Summary of frequently used notations.}
    \renewcommand\tabcolsep{3pt}
    \renewcommand\arraystretch{1.05}
    \begin{tabular}{l|l}
        \toprule[1.2pt]
         \multicolumn{1}{c|}{Notation} & \multicolumn{1}{c}{Description} \\
        \midrule
         $\mathcal{G}$& Original graph\\
         $\mathcal{G}^v$, $\mathcal{G}^m$& Masked graphs\\
         $\mathcal{E}$& Edge set\\
         $\mathcal{V}$, $\mathcal{V}^{v}$, $\mathcal{V}^{m}$ & All/visible/masked node set\\
         $\mathcal{T}^{v}$, $\mathcal{T}^{m}$& Token-masked node set of the visible/masked part\\
         $\mathcal{H}^{m}$&Token-masked representation set of the masked part\\
         $|\mathcal{V}|$, $|\mathcal{V}^{v}|$, $|\mathcal{V}^{m}|$& The number of all/visible/masked nodes\\
         $C$ &The number of data categories\\
         $D$, $d$& Node attribute/representation dimension\\
         $r$& Mask ratio\\
         $\mathbf{b} \in \mathbb{R}^{|\mathcal{V}|}$&Random binary mask vector \\
         $\mathbf{x}_i^v$ $\in \mathbb{R}^{D}$& Attribute vector of $i$-th visible node\\
         $\mathbf{x}_i^m$ $\in \mathbb{R}^{D}$&Attribute vector of $i$-th masked node\\
         $\mathbf{\widehat{x}}_i^m$ $\in \mathbb{R}^{D}$& Reconstructed attribute vector of $i$-th masked node\\
         $\widehat{\mathbf{z}}_{i}^m$ $\in \mathbb{R}^{d}$& Predicted representation vector of $i$-th masked node\\ 
         $\mathbf{z}^{m}_{i}$ $\in \mathbb{R}^{d}$& Representation vector of $i$-th masked node\\ 
         $\mathbf{t}_i^v$ $\in \mathbb{R}^{D}$& Token-masked attribute vector of $i$-th visible node\\
         $\mathbf{t}_i^m$ $\in \mathbb{R}^{D}$& Token-masked attribute vector of $i$-th masked node\\
         $\mathbf{h}_i^m$ $\in \mathbb{R}^{d}$& Token-masked representation vector of $i$-th masked node\\
         $\mathbf{\widetilde{A}} \in \mathbb{R}^{|\mathcal{V}| \times |\mathcal{V}|}$& Normalized adjacency matrix \\
         $\mathbf{X} \in \mathbb{R}^{|\mathcal{V}| \times D}$& Raw attribute matrix of all nodes\\
         $\mathbf{X}^v \in \mathbb{R}^{|\mathcal{V}^v| \times D}$&Raw attribute matrix of visible nodes\\
         $\mathbf{X}^m \in \mathbb{R}^{|\mathcal{V}^m| \times D}$&Raw attribute matrix of masked nodes\\
         $\mathbf{\widehat{X}}^m \in \mathbb{R}^{|\mathcal{V}^{m}| \times D}$& Reconstructed raw attribute matrix of masked nodes\\
         $\mathbf{T}^v \in \mathbb{R}^{|\mathcal{V}^v| \times D}$&Token-masked attribute matrix of visible nodes\\
         $\mathbf{T}^m \in \mathbb{R}^{|\mathcal{V}^m| \times D}$&Token-masked attribute matrix of masked nodes\\
         $\mathbf{H}^m \in \mathbb{R}^{|\mathcal{V}^m| \times d}$&Token-masked representation matrix of masked nodes\\
         $\mathbf{Z}^v \in \mathbb{R}^{|\mathcal{V}^{v}| \times d}$&Representation matrix of visible nodes\\
         $\mathbf{Z}^m \in \mathbb{R}^{|\mathcal{V}^{m}| \times d}$&Representation matrix of masked nodes\\
         $\mathbf{\widetilde{Z}} \in \mathbb{R}^{|\mathcal{V}| \times d}$&Recomposed representation matrix\\
         $\mathbf{\widehat{Z}}^m \in \mathbb{R}^{|\mathcal{V}^m| \times d}$&Predicted representation matrix of masked nodes\\
         $\cup$, $\cap$, $\varnothing$&Union symbol/intersection symbol/empty set\\
         $\langle\cdot,\cdot\rangle$&Inner product\\ 
        \bottomrule[1.2pt]
    \end{tabular}
    \label{I}
\end{table}

\subsection{Task Definition and Overall Framework}
\subsubsection{Task Definition}
In this study, we mainly focus on the task of self-supervised masked graph pre-training for unlabeled graphs. Our model is designed to learn two graph encoding functions (\textit{i.e.,} $\mathcal{F}_g(\cdot)$ and $\mathcal{F}_m(\cdot)$), along with a hidden predicting function (\textit{i.e.,} $\mathcal{F}_p(\cdot)$) to recover masked latent features from observations. Subsequently, a decoding function (\textit{i.e.,} $\mathcal{F}_d(\cdot)$) is employed to reconstruct the raw attributes of masked samples based on the predicted representations. The learned graph embedding can be saved and utilized for various downstream tasks, such as node classification and graph classification.

\subsubsection{Overall Framework}
\begin{figure}[!t]
\centering
\includegraphics[width=3.5in]{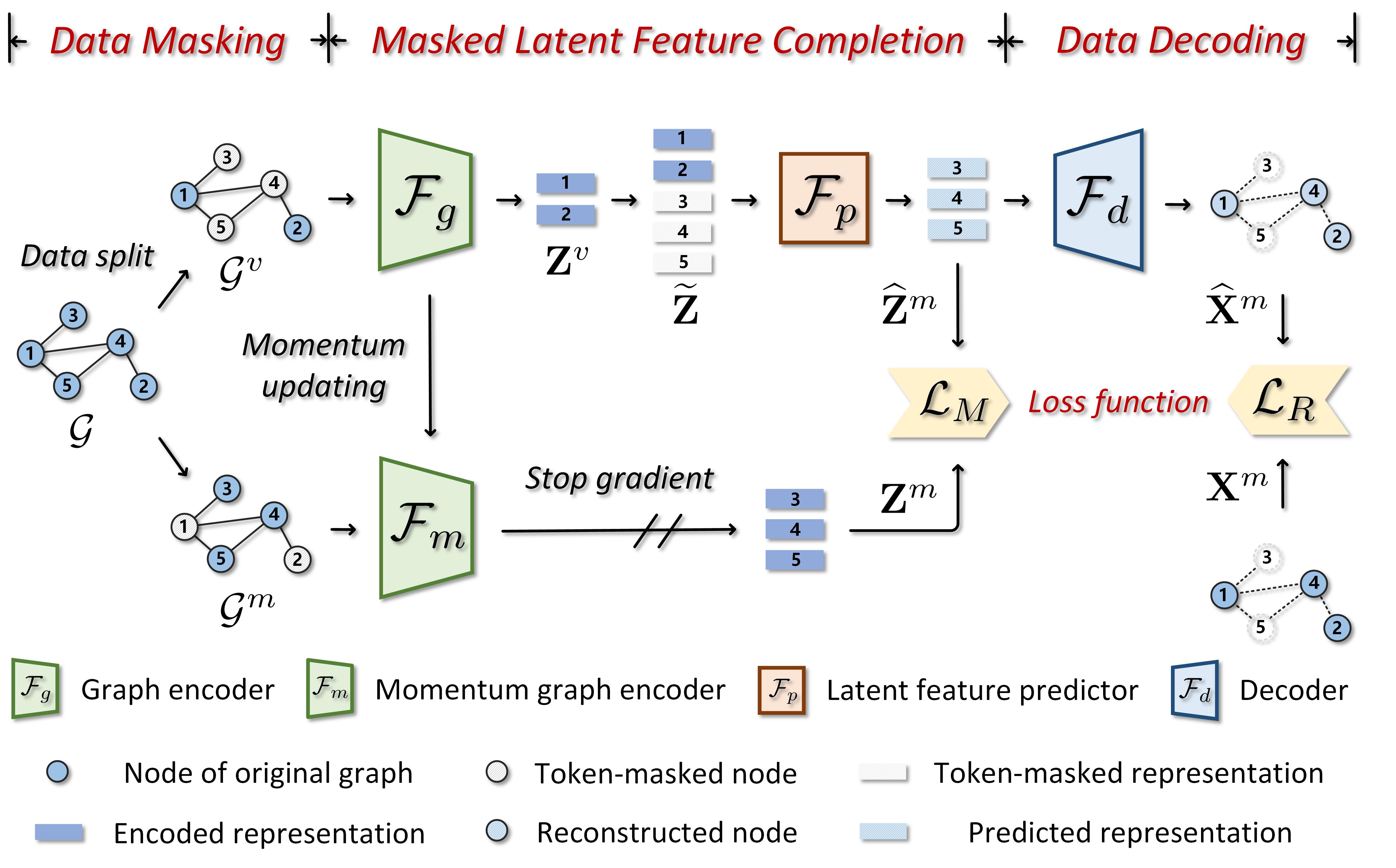}
\caption{An overview of RARE. During the pre-training phase, a graph is partitioned into two complementary masked graphs randomly. These graphs are then fed into the graph encoder and the momentum graph encoder, respectively. Next, the latent feature predictor is applied to predict the masked content from the visible one. Thereafter, the predicted representations are approximated to the output of the momentum graph encoder and processed by a simple decoder that reconstructs the raw attributes of masked nodes.}
\label{2}
\end{figure}

As shown in Fig. \ref{2}, the pre-training procedure of RARE could be mainly grouped into three parts. The goal of the data masking part is to generate two complementary masked graphs by randomly masking some nodes with tokens under a mask ratio $r$. The masked latent feature completion part is the core of RARE, which aims to enhance the reliability of the self-supervision mechanism by leveraging more informative high-order sample correlations to drive the model training. It consists of three components. Firstly, the graph encoder maps the visible nodes into node representations. Secondly, the latent feature predictor performs a latent feature prediction from visible nodes to masked ones. Thirdly, the momentum graph encoder receives the raw data of masked nodes as inputs and takes the resultant representations as implicit self-supervision signals for matching with the predicted representations. In the data decoding part, the decoder only maps the representations of masked nodes into the raw data space. Finally, $\mathcal{L}_M$ and $\mathcal{L}_R$ are integrated to minimize the loss error in both latent feature and raw data spaces. After pre-training, only the backbone of the graph encoder is adopted for downstream evaluations. The following subsections present the technical details of the corresponding components.

\subsection{Data Masking}
Before pre-training RARE, we first generate two complementary masked graphs as inputs. These graphs are then fed into the graph encoder $\mathcal{F}_g(\cdot)$ and the momentum graph encoder $\mathcal{F}_m(\cdot)$, respectively. To begin, we denote an undirected graph $\mathcal{G}=\{\mathcal{V, E}\}$ that contains $|\mathcal{V}|$ nodes with $C$ categories. Here, $\mathcal{V}$ and $\mathcal{E}$ indicate the node set and the edge set, respectively. Generally, $\mathcal{G}$ can be characterized by its normalized adjacency matrix $\mathbf{\widetilde{A}} \in \mathbb{R}^{|\mathcal{V}| \times |\mathcal{V}|}$ and raw attribute matrix $\mathbf{X} \in \mathbb{R}^{|\mathcal{V}| \times D}$, where $D$ refers to the dimension of node attributes. 
To perform the data-masking operation on a given graph, we initially draw a random binary mask vector $\mathbf{b} \in \mathbb{R}^{|\mathcal{V}|}$, where $b_i=0$ if $\mathbf{x}_i$ is masked with a node token, and $b_i=1$ otherwise. The probability of drawing 0 is $r$, which represents the mask ratio. Based on $\mathbf{b}$, we obtain the raw attribute matrix of visible nodes $\mathbf{X}^v \in \mathbb{R}^{|\mathcal{V}^v| \times D}$ and the raw attribute matrix of masked nodes $\mathbf{X}^m \in \mathbb{R}^{|\mathcal{V}^m| \times D}$:
\begin{equation} \label{eq:1}
\begin{split}
\mathbf{X}^v = \mathbf{X}[\mathbf{b}], \quad \mathbf{X}^m = \mathbf{X}[1-\mathbf{b}].
\end{split}
\end{equation} 
Accordingly, the nodes in $\mathcal{G}$ are randomly divided into two sets, \textit{i.e.,} the visible node set $\mathcal{V}^{v}=\{\mathbf{x}_i^v\}^{|\mathcal{V}^{v}|}_{i=1}$ and the masked node set $\mathcal{V}^{m}=\{\mathbf{x}_i^m\}^{|\mathcal{V}^{m}|}_{i=1}$, where $\mathcal{V}$ = $\mathcal{V}^{v}$ $\cup$ $\mathcal{V}^{m}$, $\mathcal{V}^{v}$ $\cap$ $\mathcal{V}^{m}$ = $\varnothing$. $|\mathcal{V}^v|$ and $|\mathcal{V}^m|$ denotes the numbers of visible nodes and masked nodes, respectively, where $|\mathcal{V}|$ = $|\mathcal{V}^v|$ + $|\mathcal{V}^m|$.
Similarly, we define token-masked node sets of the visible part and the masked part as $\mathcal{T}^{v}=\{\mathbf{t}_i^v\}^{|\mathcal{V}^{v}|}_{i=1}$ and $\mathcal{T}^{m}=\{\mathbf{t}_i^m\}^{|\mathcal{V}^{m}|}_{i=1}$, respectively, where $\mathbf{t}_i^v$ (or $\mathbf{t}_i^m$) $\in \mathbb{R}^{D}$ refers to a stochastic learnable vector. With these mathematical formulations, two complementary masked graphs for pre-training can be denoted as $\mathcal{G}^v=\{\mathcal{V}^v, \mathcal{T}^m, \mathcal{E}\}$ and $\mathcal{G}^m=\{\mathcal{V}^m, \mathcal{T}^v, \mathcal{E}\}$, respectively. All frequently used notations are listed in Table \ref{I}.

\subsection{Masked Latent Feature Completion} 
As discussed in the previous section, although the masked auto-encoder has proven effective and efficient, applying it to process non-Euclidean graphs directly may not always provide the required expressive capability for feature extraction. To address this issue, we propose a simple yet effective \textbf{\underline{M}}asked \textbf{\underline{L}}atent \textbf{\underline{F}}eature \textbf{\underline{C}}ompletion (MLFC) scheme. It facilitates model learning by incorporating more informative high-order sample correlations that are hard to be observed from the raw data perspective, leading to enhanced certainty in inferring masked content and a more reliable self-supervision mechanism for greater information encoding capability. The learning process of MLFC includes the following four main steps.

\subsubsection{Graph Encoding} 
The graph encoder $\mathcal{F}_g(\cdot)$ is responsible for transforming the masked graph $\mathcal{G}^v$ into a low-dimension latent space. To achieve this, we employ a graph neural network (GNN)-based architecture that consists of a sequence of graph attention layers \cite{2018GAT} or graph isomorphism layers \cite{2018GIN} as the encoder backbone. Inspired by BYOL \cite{2020BYOL}, we incorporate a multilayer perception (MLP) layer as a projector following the backbone to form the graph encoder. This encoder generates the visible node representation matrix $\mathbf{Z}^{v} \in \mathbb{R}^{|\mathcal{V}^{v}| \times d}$, where $d$ represents the latent dimension.

\subsubsection{Latent Feature Predicting} 
Following the graph encoder $\mathcal{F}_g(\cdot)$, an autoencoder-style latent feature predictor $\mathcal{F}_p(\cdot)$ is elaborately designed, which consists of two parts, \textit{i.e.,} a graph attention (or graph isomorphism) layer that recovers the masked content from observations and an MLP layer that predicts the latent features of masked nodes based on recovered information. 
Specifically, we utilize a Concat function $C(\cdot)$ to integrate $\mathbf{Z}^v$ and a token-masked representation matrix $\mathbf{H}^m \in \mathbb{R}^{|\mathcal{V}^{m}| \times d}$, where $\mathbf{h}_i^m \in \mathbb{R}^{d}$ denotes a $d$-dimensional stochastic learnable vector.
It is worth noting that the information concatenation used here to construct $\mathbf{\widetilde{Z}} \in \mathbb{R}^{|\mathcal{V}| \times d}$ is not the classic channel-wise or row-wise concatenation. Instead, we fill the visible part with $\mathbf{Z}^v$ and the masked part with $\mathbf{H}^m$ to create the recomposed representation matrix $\mathbf{\widetilde{Z}}$. Finally, we apply $\mathcal{F}_p(\cdot)$ to process $\mathbf{\widetilde{Z}}$ and obtain a predicted representation matrix of masked nodes $\mathbf{\widehat{Z}}^m \in \mathbb{R}^{|\mathcal{V}^m| \times d}$. 

\subsubsection{Momentum Graph Encoding}
Since we have obtained the predicted representations of masked nodes, a natural question arises: how to provide effective supervision to guide the masked latent feature completion in unsupervised scenarios? Our answer is to acquire the self-supervision signals from the data itself via self-distillation learning. To this end, we introduce a momentum graph encoder $\mathcal{F}_m(\cdot)$ that has the same architecture as the graph encoder. This encoder is responsible for encoding the raw data of masked samples and utilizing their complete representations to provide the predicted ones with stable optimization guidance. Concretely, we take the masked graph $\mathcal{G}^m$ as an input and feed it into $\mathcal{F}_m(\cdot)$. The resultant representation matrix $\mathbf{Z}^m \in \mathbb{R}^{|\mathcal{V}^{m}| \times d}$ preserves high-order sample correlations and serves as implicit self-supervision signals to refine $\mathbf{\widehat{Z}}^m$. 
It is worth noting that $\mathcal{F}_m(\cdot)$ is detached from the gradient back-propagation and its parameters are updated by exponential moving average (EMA) \cite{2020BYOL}. The parameters of $\mathcal{F}_g(\cdot)$ and $\mathcal{F}_m(\cdot)$ are denoted as $\Theta_{\mathcal{F}_{g}}$ and $\Theta_{\mathcal{F}_{m}}$, respectively, and the parameters of $\mathcal{F}_m(\cdot)$ are updated by:
\begin{equation} \label{eq:2}
\Theta_{\mathcal{F}_{m}} \gets \mu \Theta_{\mathcal{F}_{m}} + (1 - \mu) \Theta_{\mathcal{F}_{g}},
\end{equation}
where $\mu$ denotes the momentum factor that has been determined empirically and fixed as 0.1. Since both graph encoders involve training on multiple subsets of a common graph, the EMA can provide $\mathcal{F}_m(\cdot)$ with a smooth estimate of the underlying graph data distribution from $\mathcal{F}_g(\cdot)$, thus promoting $\mathcal{F}_m(\cdot)$ to encode reliable information. 

\subsubsection{Latent Feature Matching} 
This operation acts as an implicit form of self-supervision for recovering  masked content at the feature level. Rather than aiming to maintain similarity between the reconstructed nodes (or edges) and the raw information of the original graphs, our method focuses on ensuring that the predicted representations precisely match with the underlying structural statistics calculated by the momentum graph encoder. 
To this end, we approximate $\mathbf{\widehat{Z}}^m$ to $\mathbf{Z}^m$ by minimizing the following formulation:
\begin{equation} \label{eq:3}
\begin{split}
\mathcal{L}_M = \frac{1}{|\mathcal{V}^{m}|} \sum_{{i}=1}^{|\mathcal{V}^{m}|}\|\widehat{\mathbf{z}}^m_{i}-\mathbf{z}^m_{i}\|^{2},
\end{split}
\end{equation} 
where $\widehat{\mathbf{z}}^{m}_{i} \in \mathbb{R}^{d}$ and $\mathbf{z}^{m}_{i} \in \mathbb{R}^{d}$ indicate the representation vectors of $i$-th masked node within $\mathbf{\widehat{Z}}^m$ and $\mathbf{Z}^m$, respectively. 
According to Eq. (\ref{eq:3}), we recover masked information within the latent space by constraining the predicted representations to match with the ones that preserve more informative underlying structural information of graphs. By taking these implicit self-supervision signals as model learning guidance, we could ensure the integrity and accuracy of predicted representations, resulting in a higher quality of the learned graph embedding.

\begin{algorithm}[!t]
{\caption{Robust Masked Graph Autoencoder (RARE)}\label{Algorithm1}
\small
\begin{algorithmic}[1]
\REQUIRE Raw graph $\mathcal{G}=\{\mathcal{V, E}\}$; token-masked nodes $\{\mathcal{T}^{v}, \mathcal{T}^{m}\}$; token-masked representation matrix $\mathbf{H}^{m}$; Maximum iterations \textit{E}; mask ratio $r$; scaling factor $t$; balanced coefficient $\alpha$; model parameters $\{\Theta_{\mathcal{F}_{g}}, \Theta_{\mathcal{F}_{m}}, \Theta_{\mathcal{F}_{p}}, \Theta_{\mathcal{F}_{d}}\}$; learning rate $\eta$.
\ENSURE Pre-trained parameters $\Theta_{\mathcal{F}_{g}}$.
\STATE Initialize $\{\Theta_{\mathcal{F}_{g}}, \Theta_{\mathcal{F}_{m}}, \Theta_{\mathcal{F}_{p}}, \Theta_{\mathcal{F}_{d}}\}$ with an Xavier manner;
\FOR {$e = 1$ to $E$}
\STATE $\{\mathcal{V}^v, \mathcal{V}^m\}$ $\gets$ Split $\mathcal{V}$ into visible and masked node sets.
\STATE $\{\mathcal{G}^v, \mathcal{G}^m\}$ $\gets$ Obtain two masked graphs.
\STATE $\mathbf{Z}^v$ $\gets$ Obtain representations from $\mathcal{G}^v$ with $\mathcal{F}_g(\cdot)$.
\STATE $\mathbf{\widetilde{Z}}$ $\gets$ Integrate $\mathbf{Z}^v$ and $\mathbf{H}^m$ using $C(\cdot)$. 
\STATE $\mathbf{\widehat{Z}}$ $\gets$ Obtain predicted representations with $\mathcal{F}_p(\cdot)$.
\STATE $\mathbf{Z}^m$ $\gets$ Obtain representations from $\mathcal{G}^m$ with $\mathcal{F}_m(\cdot)$.
\STATE $\mathcal{L}_M$ $\gets$ Calculate loss error by Eq. (\ref{eq:3}).
\STATE $\mathbf{\widehat{X}}^m$ $\gets$ Obtain raw attributes from $\mathbf{\widehat{Z}}^m$ with $\mathcal{F}_d(\cdot)$.
\STATE $\mathcal{L}_R$ $\gets$ Calculate loss error by Eq. (\ref{eq:4}).
\STATE $\mathcal{L}$ $\gets$ Calculate the total loss by Eq. (\ref{eq:5}).
\STATE Update $\{\Theta_{\mathcal{F}_{g}}, \Theta_{\mathcal{F}_{p}}, \Theta_{\mathcal{F}_{d}}\}$ by calculating:\\
$\Theta_{\mathcal{F}_{g}}$ $\gets$ $\Theta_{\mathcal{F}_{g}}$ $-$ $\eta$$\nabla_{\Theta_{\mathcal{F}_{g}}}$$\mathcal{L}$;\\
$\Theta_{\mathcal{F}_{p}}$ $\gets$ $\Theta_{\mathcal{F}_{p}}$ $-$ $\eta$$\nabla_{\Theta_{\mathcal{F}_{p}}}$$\mathcal{L}$;\\
$\Theta_{\mathcal{F}_{d}}$ $\gets$ $\Theta_{\mathcal{F}_{d}}$ $-$ $\eta$$\nabla_{\Theta_{\mathcal{F}_{d}}}$$\mathcal{L}$.
\STATE Update $\Theta_{\mathcal{F}_{m}}$ by Eq. (\ref{eq:2}).\\
\ENDFOR \\
\RETURN $\Theta_{\mathcal{F}_{g}}$
\end{algorithmic}}
\end{algorithm}

\subsection{Data Decoding}
To complete the raw attributes of masked nodes, we employ a simple MLP layer as a decoder $\mathcal{F}_d(\cdot)$ to map $\mathbf{\widehat{Z}}^m$ into the raw data space. Once the reconstructed attribute matrix of masked nodes $\mathbf{\widehat{X}}^m \in \mathbb{R}^{|\mathcal{V}^{m}| \times D}$ has been processed by the decoder, we take $\mathbf{X}^m \in \mathbb{R}^{|\mathcal{V}^{m}| \times D}$ as explicit self-supervision signals and reconstruct the raw data for masked nodes by minimizing the distance between $\mathbf{\widehat{X}}^m$ and $\mathbf{X}^m$. Inspired by the SCE loss \cite{2022GraphMAE}, we design an improved scaled cosine error to boost the stability of network training, formulated as:   
\begin{equation} \label{eq:4}
\begin{split}
\mathcal{L}_R = - \frac{1}{|\mathcal{V}^{m}|}  \sum_{{i}=1}^{|\mathcal{V}^{m}|}\log\left(\frac{1}{2}+\frac{\langle \widehat{\mathbf{x}}_{i}^{m}, \mathbf{x}^{m}_{i} \rangle}{2 \|\widehat{\mathbf{x}}^{m}_{i}\| \|\mathbf{x}^{m}_{i}\|}\right)^{t},
\end{split}
\end{equation}
where $\langle\cdot,\cdot\rangle$ refers to an inner product operation. $\widehat{\mathbf{x}}_{i}^{m} \in \mathbb{R}^{D}$ and
$\mathbf{x}^{m}_{i} \in \mathbb{R}^{D}$ denotes the attribute vectors of $i$-th masked node within $\mathbf{\widehat{X}}^m$ and $\mathbf{X}^m$, respectively. $t$ is a scaling factor and we empirically set $t=2$ in most cases. 

\subsection{Loss Function and Complexity Analysis}
\subsubsection{Loss Function}
By integrating implicit and explicit self-supervision mechanisms in a united pre-training framework, the total loss of the proposed RARE can be formulated as a weighted combination of the latent feature matching loss $\mathcal{L}_M$ and the raw attribute reconstruction loss $\mathcal{L}_R$:
\begin{equation} \label{eq:5}
\begin{split}
\mathcal{L} = \mathcal{L}_M + \alpha \mathcal{L}_R,
\end{split}
\end{equation}
where $\alpha$ is a balanced coefficient. In the inference phase, the input graph $\mathcal{G}$ with $\mathbf{\widetilde{A}}$ and $\mathbf{X}$ is fed into RARE without any data-masking operations. 
The resultant graph embedding can be saved and used for downstream evaluations, such as graph classification and image recognition tasks. The detailed pre-training procedure of RARE is illustrated in Algorithm \ref{Algorithm1}.

\subsubsection{Complexity Analysis}
The time complexity of the proposed RARE could be discussed from the following two perspectives: the graph auto-encoder framework and the loss error computation. For two graph encoders, the complexities of $\mathcal{F}_g(\cdot)$ and $\mathcal{F}_m(\cdot)$ are $\mathcal{O}(LK(|\mathcal{V}|d_id_o+|\mathcal{E}|d_o))$, where $|\mathcal{V}|$, $|\mathcal{E}|$, $L$, and $K$ are the numbers of nodes, edges, encoder layers, and attention heads, respectively. $d_{i}$ and $d_{o}$ are the dimension of input and output features, respectively. For the latent feature predictor, the complexity of $\mathcal{F}_p(\cdot)$ is $\mathcal{O}(L(K(|\mathcal{V}|d_id_o+|\mathcal{E}|d_o)+|\mathcal{V}|d_id_o))$. For the decoder, the complexity of $\mathcal{F}_d(\cdot)$ is $\mathcal{O}(L|\mathcal{V}|d_id_o)$. For the computation of loss error, the time complexities of $\mathcal{L}_M$ and $\mathcal{L}_R$ are $\mathcal{O}(|\mathcal{V}^M|D)$, where $|\mathcal{V}^M|$ and $D$ are the numbers of token-masked samples and the dimension of node attributes. Therefore, the overall time complexity of RARE for each training iteration is $\mathcal{O}(L(K(|\mathcal{V}|d_id_o+|\mathcal{E}|d_o)+|\mathcal{V}|d_id_o)+|\mathcal{V}^M|D)$. We can observe that the complexity of RARE is linear with both the numbers of nodes $|\mathcal{V}|$ and edges $|\mathcal{E}|$ of the graph, making the proposed RARE theoretically efficient and scalable.

\subsection{Discussion}
In this section, we aim to explain the reasons why the proposed MLFC scheme is effective and why RARE works better than existing MGAEs, respectively. 

\subsubsection{Why The Proposed MLFC Scheme Is Effective}
We start from a more intuitive masked signal modeling perspective to revisit MLFC. As aforementioned, we can obtain the latent features of visible and masked nodes from two complementary masked views (\textit{i.e.,} $\mathcal{G}^v$ and $\mathcal{G}^m$), respectively:
\begin{equation} \label{eq:6}
\begin{split}
\mathbf{Z}^v = \mathcal{F}_g(\mathbf{X}^v, \mathbf{T}^m, \widetilde{\mathbf{A}}),
\end{split}
\end{equation}
\begin{equation} \label{eq:7}
\begin{split}
\mathbf{Z}^m = \mathcal{F}_m(\mathbf{X}^m, \mathbf{T}^v, \widetilde{\mathbf{A}}).
\end{split}
\end{equation}
After that, $\mathcal{F}_p(\cdot)$ outputs the predicted representations $\mathbf{\widehat{Z}}^m$ for masked nodes from visible ones $\mathbf{Z}^v$, and then match $\mathbf{\widehat{Z}}^m$ with $\mathbf{Z}^m$ by Eq. (\ref{eq:3}), which can be rewritten as:
\begin{equation} \label{eq:8}
\begin{split}
\mathcal{L}_M =  \mathbb{E}_{\mathbf{Z}^v,\mathbf{Z}^m}\|\mathcal{F}_p(\mathbf{Z}^v, \mathbf{H}^m, \widetilde{\mathbf{A}})-\mathbf{Z}^m\|^{2}.
\end{split}
\end{equation}
Based on Eq. (\ref{eq:8}), we observe that the MLFC scheme actually learns to pair two complementary views (\textit{i.e.,} $\mathbf{Z}^v$ and $\mathbf{Z}^m$) through a latent feature matching task. Inspired by the previous work \cite{2022U-MAE,2023UOBGE}, we denote a bipartite graph $\mathcal{G}_B = \{\mathcal{Z}^v, \mathcal{Z}^m, \mathcal{E}_B\}$ to model the corresponding learning problem, where $\mathcal{Z}^{v}=\{\mathbf{Z}^v\}^{|\mathcal{Z}^{v}|}_{i=1}$ and $\mathcal{Z}^{m}=\{\mathbf{Z}^m\}^{|\mathcal{Z}^{m}|}_{j=1}$ denotes the sets of visible views and masked views, respectively. $\mathcal{E}_B$ is represented as an adjacency matrix $\mathbf{A}_B \in \mathbb{R}^{|\mathcal{Z}^m| \times |\mathcal{Z}^v|}$ whose normalized version is $\widetilde{\mathbf{A}}_B = \mathbf{D}^{m-\frac{1}{2}}\mathbf{A}_B\mathbf{D}^{v-\frac{1}{2}}$. Here, both $\mathbf{D}^m$ and $\mathbf{D}^v$ are diagonal degree matrices. Consequently, we can derive an asymmetric instance alignment loss between $\mathbf{Z}^v$ and $\mathbf{Z}^m$ to lower bounded $\mathcal{L}_M$.

\textbf{Theorem 1.} \textit{We assume that any autoencoder-style architecture $\mathcal{F}(\cdot)$ satisfies $\mathbb{E}_\mathbf{X} \|\mathcal{F}(\mathbf{X})-\mathbf{X}\|^2 \leq \delta$, where $\mathbf{X}$ represents either visible content $\mathbf{Z}^v$ or masked content $\mathbf{Z}^m$, therefore, the MLFC loss on the bipartite graph $\mathcal{G}_B$ can be lower bounded by:}
\begin{equation} \label{eq:9}
    \begin{aligned}
    \mathcal{L}_M &\geq -\mathbb{E}_{\mathbf{Z}^v,\mathbf{Z}^m} \mathcal{F}_p(\mathbf{Z}^v, \mathbf{H}^m, \widetilde{\mathbf{A}})^\top\mathcal{F}(\mathbf{Z}^m)-\delta+1\\
    &\geq -\mathrm{tr}(\mathbf{X}^{m\top}\mathbf{\widetilde{A}}_B\mathbf{X}^{v})-\delta+1  
\end{aligned},
\end{equation}
\textit{where $\mathbf{X}^{v} \in \mathbb{R}^{|\mathcal{Z}^v| \times (|\mathcal{V}^m|d)}$ denotes the output matrix of $\mathcal{F}_{p}(\cdot)$ on $\mathcal{Z}^{v}$ whose $i$-row is $\mathbf{x}^{v}_{i}$=$\sqrt{d_i}\mathcal{F}_{p}(\mathbf{Z}^v, \mathbf{H}^m, \widetilde{\mathbf{A}})_i \in \mathbb{R}^{|\mathcal{V}^m|d}$, and $\mathbf{X}^{m} \in \mathbb{R}^{|\mathcal{Z}^m| \times (|\mathcal{V}^m|d)}$ denotes the output matrix of $\mathcal{F}(\cdot)$ on $\mathcal{Z}^{m}$ whose $j$-row is $\mathbf{x}^{m}_{j}$=$\sqrt{d_j}\mathcal{F}(\mathbf{Z}^m)_j \in \mathbb{R}^{|\mathcal{V}^m|d}$.} Note that both $\mathcal{F}_{p}(\cdot)$ and $\mathcal{F}(\cdot)$ are normalized. 

\begin{figure}[!t]
\centering
\includegraphics[width=3.2in]{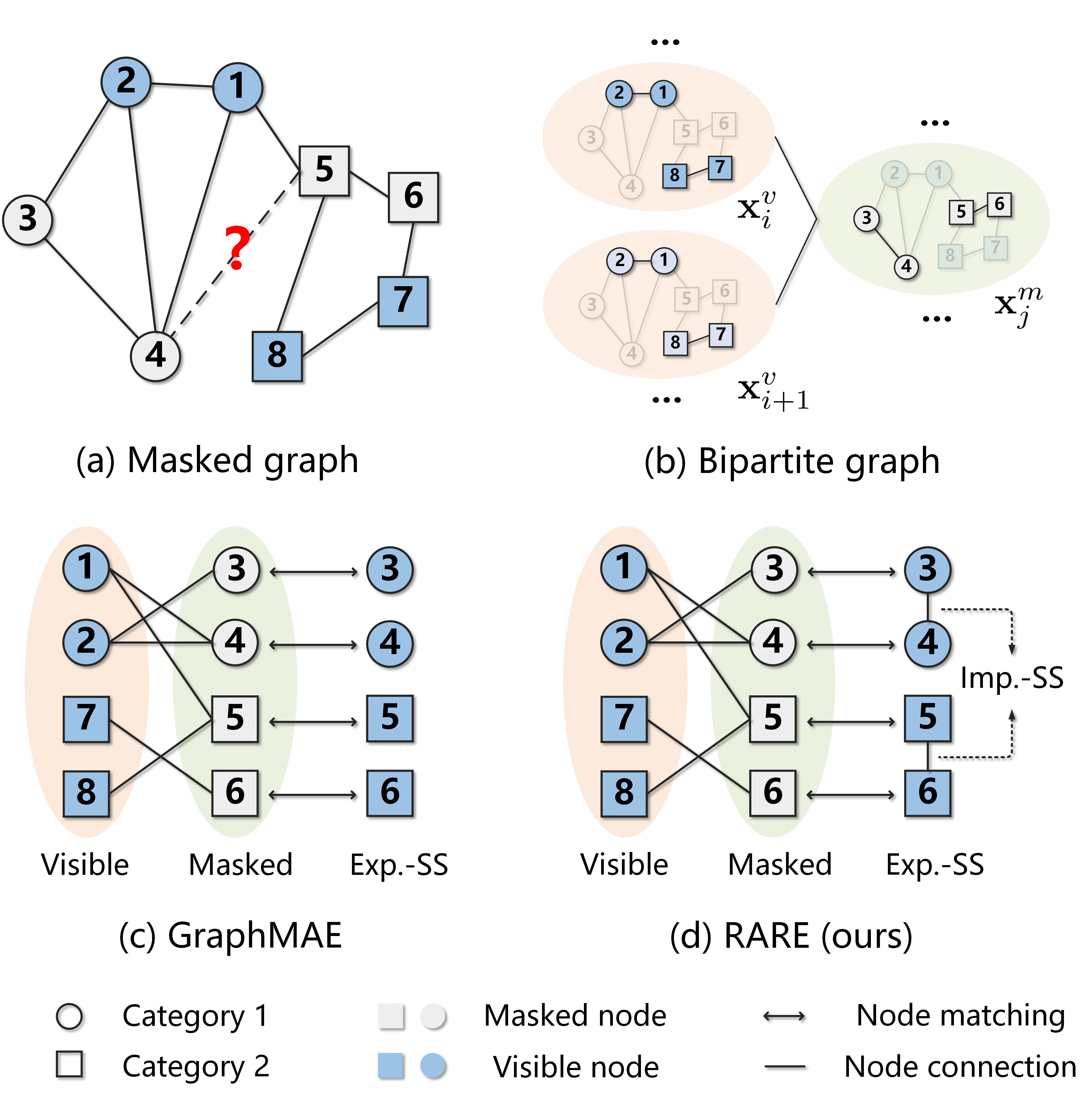}
\caption{Motivation illustration: (a) A masked graph; (b) A bipartite graph that includes two types of complementary views; (c) and (d) Comparison of self-supervision mechanisms between two methods. To drive the model training, GraphMAE \cite{2022GraphMAE} only conducts an explicit self-supervision (\textit{i.e.,} Exp.-SS) mechanism by matching the predicted nodes with raw ones, while the proposed RARE differs from GraphMAE in taking high-order sample correlations as implicit self-supervision (\textit{i.e.,} Imp.-SS) signals that are hard to be observed from the raw data perspective.}
\label{3}
\end{figure}

According to Eq. (\ref{eq:9}), it is evident that the MLFC scheme aims to minimize $\mathcal{L}_M$ by conducting an implicit similarity-based alignment between connected visible and masked samples in the latent space through an autoencoder-style predictor.
Intuitively, as illustrated in Fig. \ref{3}(b), we consider a pair of second-order neighbor visible views (\textit{e.g.,} $\mathbf{x}^{v}_{i}$ and $\mathbf{x}^{v}_{i+1}$) that share a common complementary masked view (\textit{e.g.,} $\mathbf{x}^{m}_{j}$). By enforcing $\mathbf{x}^{v}_{i}$, $\mathbf{x}^{v}_{i+1} \in \mathbf{X}^{v}$ to recover $\mathbf{x}^{m}_{j} \in \mathbf{X}^{m}$ simultaneously, the latent features of visible views are implicitly correlated through the MLFC scheme. In other words, the second-order neighbors act as positive sample pairs that are pulled closer as the instance alignment (\textit{i.e.,} positive samples should remain close in the latent space) in contrastive learning. From this perspective, the proposed MLFC scheme serves as a hidden regularization that helps encode discriminative features, thereby promoting greater information encoding capability for better downstream performance.

\subsubsection{Why RARE Works Better Than Existing MGAEs}
To better understand the superiority of RARE compared to its competitors, we conduct a comparison of the self-supervision mechanism between the proposed RARE and state-of-the-art (SOTA) GraphMAE \cite{2022GraphMAE} via a toy example illustrated in Fig. \ref{3}. 
As previously mentioned, completing masked content within a masked graph can be regarded as a complementary view pairing problem between the visible and masked data. To solve this problem, GraphMAE \cite{2022GraphMAE} adopts an explicit self-supervision (\textit{i.e.,} Exp.-SS) mechanism by matching predicted nodes with raw ones directly, which may mislead the model into a local structural ambiguity situation. For example, as shown in Fig. \ref{3}(c), two masked samples (\textit{e.g.,} Node 4 and Node 5) belonging to different categories share a common visible sample (\textit{e.g.,} Node 1). By enforcing Node 1 to reconstruct Node 4 and Node 5, GraphMAE subconsciously reduces their distance in the latent space. Consequently, the model may struggle to differentiate Node 4 and Node 5 accurately in unsupervised scenarios. This is because the self-supervision signals provided by GraphMAE only preserve raw data information that is insufficient to ascertain whether two nodes belong to the same category or not. In contrast, RARE employs both implicit and explicit self-supervision mechanisms for masked content recovery by performing a joint mask-then-reconstruct strategy in both latent feature and raw data spaces. Particularly, in RARE, the MLFC scheme could provide more informative high-order sample correlations as implicit self-supervision signals, which are not readily available in the raw data space. As shown in Fig. \ref{3}(d), by incorporating prior knowledge that 1) Node 3 or Node 6 is closely related to Node 4 or Node 5; and 2) Node 3 and Node 6 are unconnected, it would become easier for the model to infer the relationship between Node 4 and Node 5. As a result, the two nodes are pushed apart from each other in the latent space. Empirical results in Section IV-G support our claim that the proposed RARE indeed works better than GraphMAE by making the learned samples belonging to different categories more distinguishable.

\section{Experiments}
In this section, we evaluate the effectiveness of RARE against advanced SGP methods. The experiments aim to answer six research questions:
$\mathbf{RQ1:}$ How does RARE perform compared to baselines in various downstream tasks? 
$\mathbf{RQ2:}$ How does each component influence the model performance? 
$\mathbf{RQ3:}$ How is the robustness of RARE against noisy graphs?
$\mathbf{RQ4:}$ How do key hyper-parameters influence the performance of RARE?  
$\mathbf{RQ5:}$ How does the running time of RARE compare to other competitors?
$\mathbf{RQ6:}$ How do the learned node representations and reconstructed data of RARE compare to those of baselines?

In the following, we begin with a brief introduction to experimental setups, including benchmark datasets, implementation procedures, training settings, and baseline methods. Then we report experimental results with corresponding analysis.

\subsection{Evaluation Setups}
\subsubsection{Benchmark Datasets}
We conduct experiments to compare the proposed RARE with several baseline methods on seventeen datasets in total, including seven node classification datasets (\textit{i.e.,} Cora, Citeseer, Pubmed, WikiCS, Corafull, Flickr, and Yelp \cite{2019DGL}), seven graph classification datasets (\textit{i.e.,} IMDB-B, IMDB-M, PROTEINS, COLLAB, MUTAG, PTC-MR, and NCI1 \cite{2015DGK}), and three image datasets (\textit{i.e.,} Usps \cite{1990LeCun}, Mnist \cite{1998LeNet}, and Fashion-mnist \cite{2017Fashion-MNIST}). Please note that all graph datasets contain both the raw attribute matrix and the adjacency matrix, while all image datasets only have the attribute matrix. 

\subsubsection{Implementation Procedures}
The learning procedure of RARE mainly includes two steps: 1) in the pre-training task, all nodes of datasets except for Flickr and Yelp are fed into the proposed RARE for at least 20 training iterations by minimizing Eq. (\ref{eq:5}). Since Flickr and Yelp are commonly used for inductive evaluations, we follow the public data split as GraphSAINT \cite{2020GraphSAINT}, where 50\%/75\%, 25\%/10\%, and 25\%/15\% nodes are randomly sampled to form the train, validation, and test sets on Flickr and Yelp, respectively; and 2) in the downstream tasks related to nodes and images, we use the publicly available train/validation/test data split for Cora, Citeseer, Pubmed, Flickr, and Yelp. For WikiCS, Corafull, Usps, Mnist, and Fashion-mnist, since these datasets have no publicly available data split, we perform a random data split where 7\%, 7\%, and 86\% nodes are randomly sampled to form the train, validation, and test sets, respectively. We train a simple linear classifier with Adam optimizer until convergence by optimizing a cross-entropy loss by 10 times. After 10 separate runs, we report average accuracy values with standard deviations for each model. In the graph-related downstream task, as is done in GraphMAE \cite{2022GraphMAE}, we take the support vector machine as a classifier and record the results with 10-fold cross-validation after 5 separate runs. To mitigate the adverse impact of randomness, we report average accuracy (ACC) values with standard deviations for all downstream evaluations.

\begin{table*}[!t]
    \centering
    \scriptsize
    \caption{Node classification performance comparison. ``-" means unavailable source code or out-of-memory error. The \textbf{boldface} and \underline{underline} values indicate the best and the runner-up results (\%) of masked graph autoencoder methods, respectively.}
    \begin{threeparttable}
    \renewcommand\tabcolsep{8pt}
    \renewcommand\arraystretch{1.05}
    \begin{tabular}{c|c|ccccccc}
        \toprule[1.2pt]
         Learning Type  & Method &   Cora      & Citeseer      & Pubmed     & WikiCS               & Corafull   &Flickr&  Yelp   \\
         \midrule
        \multirow{2}{*}{Supervised} 
        & GCN \cite{2017GCN}    &  81.5          & 70.3          & 79.0                      & 66.7$\pm$0.5& 48.6$\pm$0.5 & 42.9$\pm$0.1 & 57.3$\pm$0.1  \\
        & GAT \cite{2018GAT}    &  83.0$\pm$0.7  & 72.5$\pm$0.7  & 79.0$\pm$0.3             & 69.4$\pm$1.0 & 50.7$\pm$0.2 &  43.9$\pm$0.1& 57.6$\pm$0.1      \\
        \midrule
        \multirow{10}{*}{Self-supervised} 
        & GAE \cite{2016VGAE}    &  71.5$\pm$0.4  & 65.8$\pm$0.4  & 72.1$\pm$0.5            &67.3$\pm$0.3 & 52.0$\pm$0.1 & -& -\\

        & DGI \cite{2019DGI}    &  82.3$\pm$0.6  & 71.8$\pm$0.7  & 76.8$\pm$0.6           & 64.8$\pm$0.6 & 48.2$\pm$0.5 & 45.0$\pm$0.2 & 57.4$\pm$0.1 \\
        & MVGRL \cite{2020MVGRL}  & 83.5$\pm$0.4   & 73.3$\pm$0.5  & 80.1$\pm$0.7                      & 64.8$\pm$0.7 & 52.6$\pm$0.5 & -&-\\
        & GRACE \cite{2020GRACE}  & 81.9$\pm$0.4   & 71.2$\pm$0.5  & 80.6$\pm$0.4          & 68.0$\pm$0.7 &45.2$\pm$0.1 & -&-\\  
        & BGRL \cite{2021BGRL}  & 82.7$\pm$0.6   & 71.1$\pm$0.8  & 79.6$\pm$0.5            &65.5$\pm$1.5 & 47.4$\pm$0.5& 39.4$\pm$0.1&-\\
        & InfoGCL \cite{2021InfoGCL} & 83.5$\pm$0.3   & 73.5$\pm$0.4  & 79.1$\pm$0.2 &-& -&- &-\\
        & CCA-SSG\cite{2021CCA--SSG} & 84.0$\pm$0.4  & 73.1$\pm$0.3  & 81.0$\pm$0.4   &67.4$\pm$0.9 &53.5$\pm$0.4 &49.1$\pm$0.1 &59.6$\pm$0.1 \\
        & SeeGera \cite{2023SeeGera} & 82.8$\pm$0.3& 71.6$\pm$0.2 &  79.2$\pm$0.3&     65.8$\pm$0.2   &       -   &     -     &-\\
        & MaskGAE \cite{2022MaskGAE} & 82.6$\pm$0.3 &  \underline{73.4$\pm$0.6} & 81.0$\pm$0.3 &  \underline{66.0$\pm$0.2}  & 52.2$\pm$0.1 &49.1$\pm$0.1& 68.1$\pm$0.1  \\
        &  GraphMAE \cite{2022GraphMAE}&  \textbf{84.2$\pm$0.4}  &  73.4$\pm$0.4  &  \underline{81.1$\pm$0.4}   &65.7$\pm$0.7 & \underline{53.4$\pm$0.1}& \underline{49.6$\pm$0.2}& \underline{69.4$\pm$0.2} \\
        \cmidrule{2-9}
         & Ours  & \textbf{84.2$\pm$0.2} & \textbf{74.1$\pm$0.3} & \textbf{81.8$\pm$0.2} &  \textbf{69.0$\pm$0.6} & \textbf{55.5$\pm$0.1} & \textbf{50.6$\pm$0.1} & \textbf{72.1$\pm$0.6}   \\
        \bottomrule[1.2pt]
    \end{tabular}
    \end{threeparttable}

    \label{II}
\end{table*}

\begin{table*}[!t]
    \centering
    \scriptsize
    \caption{Graph classification performance comparison. ``-" means unavailable source code or out-of-memory error. The \textbf{boldface} and \underline{underline} values indicate the best and the runner-up results (\%) of masked graph autoencoder methods, respectively.}
    \renewcommand\arraystretch{1.05}
    \begin{tabular}{c|c|ccccccc}
        \toprule[1.2pt]
          Learning Type    & Method   & IMDB-B              & IMDB-M              & PROTEINS            & COLLAB              & MUTAG               &    PTC-MR      & NCI1  \\
        \midrule
        \multirow{2}{*}{Supervised}
        & GIN \cite{2018GIN} & 75.1$\pm$5.1  & 52.3$\pm$2.8   & 76.2$\pm$2.8   & 80.2$\pm$1.9   & 89.4$\pm$5.6 & 63.7$\pm$8.2 & 82.7$\pm$1.7 \\
        & DiffPool \cite{2018DIFFPOOL} & 72.6$\pm$3.9    & -    & 75.1$\pm$3.5  & 78.9$\pm$2.3    & 85.0$\pm$10.3  &  -  & -                 \\
        \midrule
        \multirow{2}{*}{Graph Kernels}
        & WL \cite{2011Weisfeiler-Lehman}       & 72.3$\pm$3.4          & 47.0$\pm$0.5          & 72.9$\pm$0.6          & -                   & 80.7$\pm$3.0          &    58.0$\pm$0.5     & 80.3$\pm$0.5        \\
        & DGK \cite{2015DGK}     & 67.0$\pm$0.6 & 44.6$\pm$0.5 & 73.3$\pm$0.8    & -    & 87.4$\pm$2.7    & 60.1$\pm$2.6 & 80.3$\pm$0.5        \\
        \midrule
        \multirow{9}{*}{Self-supervised}
        & Graph2vec \cite{2017graph2vec}& 71.1$\pm$0.5          & 50.4$\pm$0.9         & 73.3$\pm$2.1          & -                   & 83.2$\pm$9.3          &    60.2$\pm$6.9    & 73.2$\pm$1.8        \\
        & Infograph \cite{2020Infograph}& 73.0$\pm$0.9          & 49.7$\pm$0.5          & 74.4$\pm$0.3          & 70.7$\pm$1.1          & 89.0$\pm$1.1          &    61.7$\pm$6.4     & 76.2$\pm$1.1        \\
        & GraphCL \cite{2020GraphCL}  & 71.1$\pm$0.4          & 48.6$\pm$0.7          & 74.4$\pm$0.5          & 71.4$\pm$1.2          & 86.8$\pm$1.3          &     -   & 77.9$\pm$0.4        \\
        & JOAO \cite{2021JOAO}    & 70.2$\pm$3.1          & 49.2$\pm$0.8          & 74.6$\pm$0.4          & 69.5$\pm$0.3          & 87.4$\pm$1.0          &  -    & 78.1$\pm$0.5        \\
        & GCC \cite{2020GCC}      & 72.0                & 49.4                & -                   & 78.9                & -                   &       -       & -                 \\
        & MVGRL \cite{2020MVGRL}    & 74.2$\pm$0.7          & 51.2$\pm$0.5          & -                   & -                   &  89.7$\pm$1.1    &   62.5$\pm$1.7      & -                 \\
        & InfoGCL \cite{2021InfoGCL} & 75.1$\pm$0.9          & 51.4$\pm$0.8          & -                   & 80.0$\pm$1.3          & 91.2$\pm$1.3 &         59.4$\pm$1.6     & 80.2$\pm$0.6        \\
        & GraphMAE \cite{2022GraphMAE} & \underline{75.5$\pm$0.7}    & \underline{51.6$\pm$0.5}         & \underline{75.3$\pm$0.4}         & \underline{80.3$\pm$0.5}    & \underline{88.2$\pm$1.3}          &   \underline{57.6$\pm$0.8}  & \underline{80.4$\pm$0.3}        \\
        \cmidrule{2-9}
        & Ours      &  \textbf{76.2$\pm$0.1} & \textbf{53.1$\pm$0.1} & \textbf{76.4$\pm$0.2} & \textbf{81.2$\pm$0.6} & \textbf{88.6$\pm$0.6} & \textbf{59.3$\pm$0.7} & \textbf{81.3$\pm$0.4} \\
        \bottomrule[1.2pt]
    \end{tabular}
    \label{III}
\end{table*}
\subsubsection{Training Settings}
To ensure a fair comparison, all experiments are conducted on the same device and under an identical configuration environment. For all compared baselines, we directly report the results listed in the existing literature if available. Otherwise, we implement their official source codes and report the reproduced performance. 
For our method, we perform a grid search to select hyper-parameters on the following searching space: the mask ratio $r$ is selected between \{0.5, 0.75\}; the balanced coefficient $\alpha$ is searched from 1 to 9; the scale factor $t$ is selected between \{1, 2\}; the hidden size of latent features is selected from \{256, 512, 1024\}; the number of feature extractor layers is selected from \{1, 2, 3, 4, 5\}; by default, the momentum rate is empirically fixed to 0.1 by default; the learning rate of the Adam optimizer is selected from \{1.5e-4, 5e-4, 1e-3\}; the maximum epoch is determined according to the results of model convergence. Particularly, in the graph classification task, we 1) choose the batch size from \{32, 64\}, similar to GraphMAE \cite{2022GraphMAE}; 2) consistently adopt a batch normalization operation to regularize the model learning; and 3) follow the sample pooing setups in GraphMAE, where a non-parameterized graph pooling function (\textit{e.g.,} max-pooling, mean-pooing or sum-pooling) is employed to generate graph-level representations. Please note that we employ similar hyper-parameter setups as reported in GraphMAE \cite{2022GraphMAE}, and most hyper-parameters are not carefully tuned for ease of model learning. 

\subsubsection{Baseline Methods}
In this work, we compare the performance of the proposed RARE with several baseline methods on three different tasks: node classification, graph classification, and image classification. 
For node classification, we compare RARE with two supervised methods (\textit{i.e.,} GCN \cite{2017GCN} and GAT \cite{2018GAT}) and ten self-supervised methods (\textit{i.e.,} GAE \cite{2016VGAE}, DGI \cite{2019DGI}, MVGRL \cite{2020MVGRL}, GRACE \cite{2020GRACE}, BGRL \cite{2021BGRL}, InfoGCL \cite{2021InfoGCL}, CCA-SSG \cite{2021CCA--SSG}, SeeGera \cite{2023SeeGera}, MaskGAE \cite{2022MaskGAE}, and GraphMAE \cite{2022GraphMAE}). 
For graph classification, we consider twelve baseline methods from three different aspects: 1) two supervised methods (\textit{i.e.,} GIN \cite{2018GIN} and DiffPool \cite{2018DIFFPOOL}); 2) two graph kernels-based methods (\textit{i.e.,} WL \cite{2011Weisfeiler-Lehman} and DGK \cite{2015DGK}); and 3) eight self-supervised methods (\textit{i.e.,} Graph2vec \cite{2017graph2vec}, Infograph \cite{2020Infograph}, GraphCL \cite{2020GraphCL}, JOAO \cite{2021JOAO}, GCC \cite{2020GCC}, MVGRL \cite{2020MVGRL}, InfoGCL \cite{2021InfoGCL}, and GraphMAE \cite{2022GraphMAE}). For image classification, we investigate two image classification methods (\textit{i.e.,} VGG16 \cite{2015VGG16} and ResNet18 \cite{2016ResNet18}) and three graph autoencoder methods (\textit{i.e.,} GAE \cite{2016VGAE}, MaskGAE \cite{2022MaskGAE}, and GraphMAE \cite{2022GraphMAE}.

\subsection{Overall Performance (RQ1)}
\subsubsection{Evaluation on Node Classification}
As shown in Table \ref{II}, we report the node classification performance of thirteen compared methods on seven datasets. From these results, several major observations can be concluded: 1) the proposed RARE consistently outperforms two supervised methods on all datasets, with margins going up to 7.7\%-14.8\% on Flickr and Yelp. These improvements demonstrate the great potential of masked graph autoencoders for effectively handling massive unlabeled graph data; 2) InfoGCL is one of the strongest contrastive self-supervised methods, while the proposed RARE improves it by 0.7\%, 0.6\%, 2.7\% accuracy on Cora, Citeseer, and Pubmed, respectively. This phenomenon indicates that RARE can effectively boost the learned representations by conducting a mask-then-reconstruct mechanism instead of relying on a relatively complicated contrastive mechanism; 3) taking the results on WikiCS for example, RARE significantly outperforms SeeGera, MaskGAE, and GraphMAE by 3.2\%, 3.0\%, and 3.3\%, respectively. These benefits are attributed to the novel idea of integrating implicit and explicit self-supervision mechanisms to drive model learning by performing a joint mask-then-reconstruct strategy in both latent feature and raw data spaces; and 4) on Cora, RARE achieves competitive results compared to the most powerful masked graph autoencoder, \textit{i.e.,} GraphMAE. However, it is possible that the full potential of model optimization was not demonstrated due to the insufficient size of the test dataset. Increasing the size of the test dataset, such as WikiCS and Yelp, could reveal that RARE has the potential to yield even more substantial performance gains.  

\subsubsection{Evaluation on Graph Classification}
Table \ref{III} summarizes graph classification results of thirteen methods on seven datasets. The results reveal several key observations that are similar to those obtained from the node classification task: 1) the performance of RARE is highly competitive compared to both supervised methods and graph kernels methods, indicating that the masked graph autoencoder has the potential to be a promising alternative for self-supervised graph pre-training; 2) compared to GraphCL and JOAO, our method achieves significant performance improvements (up to 1.8\%-11.7\%) over them on almost all datasets. However, some contrastive learning methods, such as InfoGCL and MVGRL, demonstrate better performance than masked graph autoencoders on MUTAG and PTC-MR. This may be because in some cases, the partitioning of small-scale graph data can be easily achieved through multi-view contrastive learning; and 3) RARE achieves an approximate 1.1\% average performance gain over GraphMAE on all datasets, which further indicates that RARE can effectively leverage both implicit and explicit self-supervision signals to improve the quality of the learned graph embedding.

\begin{table}[!t]
    \centering
    \scriptsize
    \caption{Image classification performance comparison. ``- " means the out-of-memory error. The \textbf{boldface} and \underline{underline} values indicate the best and the runner-up results (\%) of masked graph autoencoder methods, respectively. }
    \renewcommand\tabcolsep{4.5pt}
    \renewcommand\arraystretch{1.05}
    \begin{tabular}{c|c|cccc}
        \toprule[1.2pt]
         Learning Type  & Method    & Usps    & Mnist    & Fashion-mnist \\
        \midrule
        \multirow{2}{*}{Supervised}& VGG16 \cite{2015VGG16}     &  94.5$\pm$0.5   & 95.6$\pm$0.8  & 85.6$\pm$0.6  \\
        & ResNet18 \cite{2016ResNet18}  &  94.2$\pm$1.6   &  96.1$\pm$1.3  & 85.4 $\pm$1.5   \\
         \midrule
       \multirow{4}{*}{Self-supervised} & GAE  \cite{2016VGAE}   &  75.8$\pm$0.2  &  - & -  \\
        & MaskGAE \cite{2022MaskGAE}   &  92.2$\pm$0.2   &  87.1$\pm$0.1  & 78.5$\pm$0.1 \\ 
        & GraphMAE \cite{2022GraphMAE} &  \underline{93.0$\pm$0.5}  &  \underline{91.7$\pm$0.1}   & \underline{79.8$\pm$0.2} \\
        \cmidrule{2-5}
        & Ours    &  \textbf{94.3$\pm$0.3}  & \textbf{94.2$\pm$0.2} & \textbf{85.7$\pm$0.1}  \\  
        \bottomrule[1.2pt]
    \end{tabular}
    \label{IV}
\end{table}

\subsubsection{Evaluation on Image Classification}
To verify the superiority of RARE in-depth, Table \ref{IV} reports the image classification performance of six methods on three datasets. From those results, we can obtain the following observations: 1) the proposed RARE shows a significant advantage against existing state-of-the-art MGAEs and other baselines on all image benchmarks. For example, on Mnist and Fashion-mnist datasets, RARE consistently outperforms the best edge-masking-based MaskGAE and node-masking-based GraphMAE by 7.1\%/2.5\% and 7.2\%/5.9\% accuracy, respectively. These improvements once again demonstrate the effectiveness of introducing implicit self-supervision signals for model learning; and 2) it is interesting to note that RARE can achieve competitive or slightly better results than typical supervised classification methods, such as VGG16 and ResNet18. This implies that improving the reliability of the self-supervision mechanism can facilitate RARE to unleash its potential for SGP. Thus, the learned representations show good robustness and generalization across a wide range of downstream tasks.

\subsection{Ablation Study (RQ2)}
\subsubsection{Impact of The MLFC Scheme} 
To demonstrate the effectiveness of the proposed masked latent feature completion scheme, we compare RARE with its three variants on eight datasets. Concretely, ``w/o-Pred." implies that RARE removes the latent feature predictor. ``w/o-Mome." indicates that RARE discards the momentum graph encoder. ``w-$\mathcal{F}_m$($\mathcal{G}$)" denotes that the momentum graph encoder of RARE accepts $\mathcal{G}$ rather than $\mathcal{G}^m$. As shown in Table \ref{V}, some major observations can be summarized: 1) when compared to ``w/o-Pred.", the latent feature predictor produces a performance gain of 1.4\%-5.2\% on eight datasets, indicating that this component plays a vital role in our SGP solution. By iteratively conducting the mask-then-reconstruct operation on incomplete graphs, the latent feature predictor could be regarded as a hidden regularization that assists the graph encoder in extracting more compressed features; 2) RARE consistently outperforms ``w/o-Mome." on all eight datasets. Taking the results on Corafull and IMDB-M for example, RARE achieves 3.6\% and 2.3\% accuracy gains respectively, demonstrating the effectiveness of providing implicit self-supervision signals to ensure the integrity and accuracy of predicted representations. Similar observations can be concluded from the results on other datasets; and 3) although ``w-$\mathcal{F}_m$($\mathcal{G}$)" can also achieve competitive performance, it suffers from 0.3\%-1.3\% accuracy degradation compared to our method. The reason behind this may be that since $\mathcal{G}^v$ is a sub-graph of the original graph $\mathcal{G}$, a large amount of redundant information between two-source encoded representations would overwhelm the latent space, resulting in inferior representations for downstream tasks.

\begin{table}[!t]
    \centering
     \caption{Ablation study for the MLFC scheme. ``w/o-Pred." and ``w/o-Mome." denote two RARE variants with the latent feature predictor and the momentum graph encoder being removed, respectively. ``w-$\mathcal{F}_m$($\mathcal{G}$)" is a variant of RARE whose momentum graph encoder accepts a complete graph $\mathcal{G}$. $\downarrow$ denotes the performance degradation. The \textbf{boldface} values indicate the best results (\%).}
    \scriptsize
    \renewcommand\tabcolsep{5.2pt}
    \renewcommand\arraystretch{1.05}
    \begin{tabular}{c|ccccc}
        \toprule[1.2pt]
          Dataset  & w/o-Pred. &  w/o-Mome. &  w-$\mathcal{F}_m$($\mathcal{G}$)  & Ours \\
        \midrule
       Cora & 82.5 (1.7 $\downarrow$) & 83.4 (0.8 $\downarrow$)&  83.7 (0.5 $\downarrow$)&$\textbf{84.2}$ \\
        Citeseer&72.7 (1.4 $\downarrow$)&  70.5 (3.6 $\downarrow$) &  73.6 (0.5 $\downarrow$)& $\textbf{74.1}$  \\
        Pubmed & 79.5 (2.3 $\downarrow$)& 78.8 (3.0 $\downarrow$)& 81.4 (0.4 $\downarrow$)& $\textbf{81.8}$ \\
     Corafull & 52.2 (3.3 $\downarrow$)&51.9 (3.6 $\downarrow$) & 55.2 (0.3 $\downarrow$) & $\textbf{55.5}$\\  
     \midrule
      IMDB-B &74.6 (1.6 $\downarrow$)&74.7 (1.5 $\downarrow$)& 75.6 (0.6 $\downarrow$) &$\textbf{76.2}$ \\
      IMDB-M &51.7 (1.4 $\downarrow$) &50.8 (2.3 $\downarrow$)& 52.3 (0.8 $\downarrow$) &$\textbf{53.1}$ \\
      PROTEINS & 74.1 (2.3 $\downarrow$)&75.6 (0.8 $\downarrow$) & 75.4 (1.0 $\downarrow$) & $\textbf{76.4}$\\
      PTC-MR &  54.1 (5.2 $\downarrow$)  & 57.5 (1.8 $\downarrow$)   & 58.0 (1.3 $\downarrow$)  &  $\textbf{59.3}$\\  
        \bottomrule[1.2pt]
    \end{tabular}
    \label{V}
\end{table}

\begin{table}[!t]
    \centering
     \caption{Ablation study for loss functions $\mathcal{L}_M$ and $\mathcal{L}_R$. MAE, MSE, and ISCE are abbreviations for mean absolute error, mean square error, and improved scaled cosine error, respectively. The \textbf{boldface} values indicate the best results (\%).}
    \scriptsize
    \renewcommand\tabcolsep{3pt}
    \renewcommand\arraystretch{1.05}
    \begin{tabular}{c|c|cccc}
        \toprule[1.2pt]
         Method & Loss Function   &  WikiCS   & Flickr    & IMDB-M  &MUTAG\\
        \midrule
         (A) &  $\mathcal{L}_M$ (MSE) \& $ \mathcal{L}_R$ (MSE)    &  66.2   &  49.7  & 51.0 & 85.7\\
         (B)  & $\mathcal{L}_M$ (ISCE) \& $ \mathcal{L}_R$ (MSE)    & 66.5 &  49.8 & 50.0&  87.3  \\
         (C) & $\mathcal{L}_M$ (ISCE) \& $ \mathcal{L}_R$ (ISCE)  &  68.2& 49.2 & 51.4& 87.2 \\
         (D)  & $\mathcal{L}_M$ (MAE) \& $ \mathcal{L}_R$ (ISCE) &  68.8  & 49.9  & $\textbf{53.2}$ &  87.6   \\
          \midrule
         Ours  & $\mathcal{L}_M$ (MSE) \& $ \mathcal{L}_R$ (ISCE)    &  $\textbf{69.0} $  & $\textbf{50.6} $ &  53.1 & $\textbf{88.6}$\\
        \bottomrule[1.2pt]
    \end{tabular}
    \label{VI}
\end{table}

\subsubsection{Impact of The Loss Function}
In this subsection, we conduct ablation studies to investigate the effect of different loss functions. Table \ref{VI} reports the accuracy results of RARE and its four variants on WikiCS, Corafull, IMDB-M, and MUTAG. $\mathcal{L}_M$ and $\mathcal{L}_R$ indicate the loss functions used for implicit and explicit self-supervision mechanisms, respectively. Moreover, MAE, MSE, and ISCE are abbreviations for mean absolute error, mean square error, and improved scaled cosine error, respectively. From the results presented in Table \ref{VI}, we can observe that 1) our method achieves better performance than method (A) and method (B) by 2.8\%/2.5\% and 2.1\%/3.1\% accuracy improvements on WikiCS and IMDB-M, respectively. The reason behind this is that the MSE loss is better at modeling detailed information from the data itself, while the ISCE loss focuses more on estimating the similarity between two entities. Therefore, minimizing MSE in the noisy raw data space may mislead the network to overly preserve redundant graph details, which may not always result in the required expressive encoding capability for downstream tasks; and 2) RARE and method (D) consistently outperform method (C) by 0.8\%/0.6\%, 1.4\%/0.7\%, 1.7\%/1.8\%, and 1.4\%/0.4\% on WikiCS, Flickr, IMDB-M, and MUTAG, respectively. This is because the latent features contain much category-related information that needs to be carefully preserved. As a result, completing the masked content in the latent space with MAE or MSE contributes more to the performance than that with ISCE. Moreover, when one has to choose a loss function for masked latent feature completion, both MAE and MSE are suitable for guiding the model learning implicitly.  

\begin{figure*}[!t]
\centering
\includegraphics[width=7.2in]{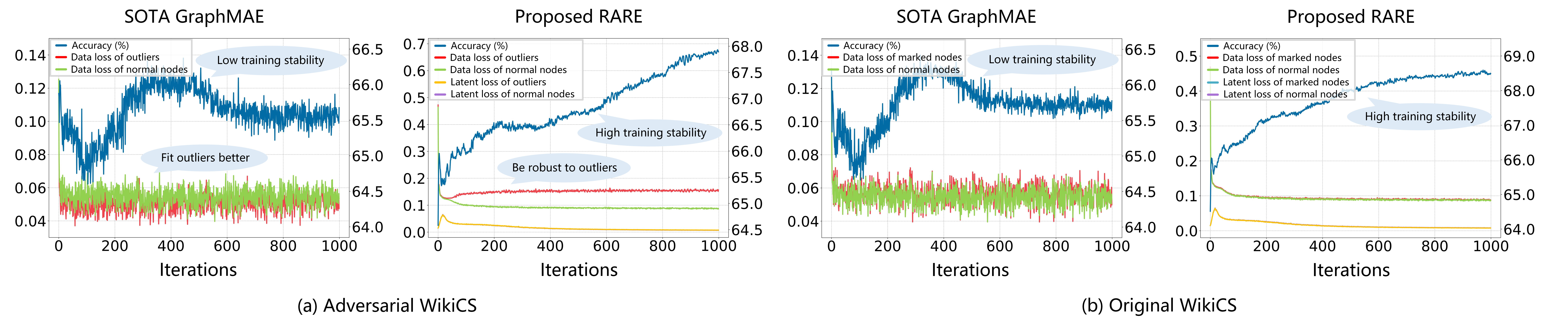}
\caption{Method robustness comparison between the proposed RARE and SOTA GraphMAE \cite{2022GraphMAE}: (a) On the adversarial WikiCS; (b) On the original WikiCS. Note that the corresponding results on the original WikiCS are provided as a reference. In our settings, both accuracy and loss variations are recorded with iterations. Before pre-training, we randomly generate 5\% outliers (\textit{i.e.,} noisy samples) from the original WikiCS by row-wise shuffling their features within the raw attribute matrix. During pre-training, a large random subset of graph nodes (\textit{e.g.,} 75\%) is masked out at first and then recovered by minimizing loss values of masked nodes only. After pre-training, we evaluate the model by reporting the performance and the loss curves of normal nodes and all outliers, respectively. As seen, RARE delivers better accuracy and is more robust than GraphMAE \cite{2022GraphMAE} in a noisy circumstance.}
\label{5}
\end{figure*}

\begin{table}[!t]
    \centering
     \caption{Ablation study for data-masking setups. ``M-Node (Z)" and ``M-Rep. (Z)" means that nodes and representations are masked with zero values, and ``M-Node (T)" and ``M-Rep. (T)" means that nodes and representations are masked with tokens (\textit{i.e.,} stochastic learnable vectors). ``w/o M-Rep." denotes a variant of RARE without masking representations. The \textbf{boldface} values indicate the best results (\%).}
    \scriptsize
    \renewcommand\tabcolsep{4.0pt}
    \renewcommand\arraystretch{1.05}
    \begin{tabular}{c|c|cccc}
        \toprule[1.2pt]
         Method & Data-masking Setup & Pubmed  & Flickr & IMBD-M  & NCI1 \\
        \midrule
         (A) & M-Node (Z)  \&  M-Rep. (Z)    & 81.2 & 49.6 & 51.7 & 80.2 \\
         (B) & M-Node (Z)  \&  M-Rep. (T)    & 80.4 & 49.4 & 52.0 & 80.6 \\
         (C) & M-Node (T)  \&   M-Rep. (Z)   & 81.2 & 49.3 & 52.5 & 80.0 \\
         (D) & M-Node (T)  \&   w/o M-Rep.   & 80.0 & 49.0 & 52.2 & 79.5 \\
          \midrule
         Ours  & M-Node (T) \&  M-Rep. (T)    &  $\textbf{81.8} $  & $\textbf{50.6} $ &  $\textbf{53.1}$ & $\textbf{81.3}$\\
        \bottomrule[1.2pt]
    \end{tabular}
    \label{VII}
\end{table}

\begin{table}[!t]
    \centering
    \caption{Ablation study for the latent feature predictor setups. ``GCL", ``GAL", ``GIL", and ``MLP" denote the graph convolution layer, graph attention layer, graph isomorphism layer, and multilayer perception, respectively. The \textbf{boldface} values indicate the best results (\%).}
    \scriptsize
    \renewcommand\tabcolsep{3.0pt}
    \renewcommand\arraystretch{1.05}
    \begin{tabular}{c|c|cccc}
        \toprule[1.2pt]
         Method & Predictor Setup &  Pubmed   & Corafull    & PROTEINS  & PTC-MR\\
        \midrule
         (A) &  GCL        & 78.8 & 49.9 & 74.6 &  56.6 \\
         (B)  & GAL (or GIL)  & 80.2 & 50.9 & 75.3 &  57.5\\
         (C) &  GCL + MLP  & 80.9 & 54.7 & 75.2 &  57.8 \\
         \midrule
         Ours & GAL (or GIL) + MLP  & \textbf{81.8} & \textbf{55.5} & \textbf{76.4} &  \textbf{59.3} \\
        \bottomrule[1.2pt]
    \end{tabular}
    \label{VIII}
\end{table}

\subsubsection{Impact of Data-masking Setups}
To investigate the effect of different data-masking setups, we conduct ablation studies to make a comparison among RARE and its four variants on Pubmed, Flickr, IMBD-M, and NCI1. Specifically, ``M-Node (Z)" and ``M-Rep. (Z)" means that nodes and representations are masked with zero values, and ``M-Node (T)" and ``M-Rep. (T)" means that nodes and representations are masked with tokens (\textit{i.e.,} stochastic learnable vectors). ``w/o M-Rep." denotes a variant of RARE without masking representations. Some major conclusions can be summarized from the results in Table \ref{VII}: 1) our method consistently achieves better accuracy performance than method (A), method (B), and method (C) on four datasets, indicating that using parameterized tokens for masked samples as initialization is more effective; and 2) the last two rows reveal the advantage of implementing the data-masking operation over latent features. It is worth noting that method (D) refines the encoded representations of masked nodes via latent feature matching directly, while RARE first recovers the masked representations from scratch via a latent feature predictor and then conducts the sample matching in the high-order latent feature space. The latter makes the self-supervised task more challenging, which further encourages the model to enhance its local modeling capability. As seen, RARE consistently outperforms method (D) by 1.8\%, 1.6\%, 0.9\%, and 1.8\% accuracy on all datasets.

\subsubsection{Impact of Latent Feature Predictor Setups}
To investigate how different latent feature predictor setups influence the performance of RARE, we compare RARE with its three variants and report their performance on Pubmed, Corafull, PROTEINS, and PTC-MR in Table \ref{VIII}. The abbreviations used in the table are as follows: ``GCL", ``GAL", ``GIL", and ``MLP" denote the graph convolution layer, graph attention layer, graph isomorphism layer, and multilayer perception, respectively. Firstly, we observe that our method produces better performance than method (C) on all datasets, indicating that GAL (or GIL) is a better option than GCL for recovering masked latent features due to its more powerful graph modeling capability. Similar observations can be obtained from the comparison between method (A) and method (B). Secondly, we also observe that our method consistently outperforms method (B) by 1.6\%, 4.6\%, 1.1\%, and 1.8\% on all datasets. Similarly, method (C) performs better than method (A) in almost all cases. These results demonstrate that MLP plays an important role in predicting the latent features of masked nodes in the MLFC scheme.

\subsection{Robustness Against Outliers (RQ3)}
To provide a more comprehensive understanding of our motivations, we conduct an experiment to make a comparison between the proposed RARE and state-of-the-art (SOTA) GraphMAE on adversarial WikiCS. We also include results from the original WikiCS as a reference. In our settings, we randomly divide all nodes of the original WikiCS into 5\% marked nodes and 95\% normal nodes. To construct an adversarial WikiCS, within the raw attribute matrix, we row-wise shuffle the attributes of 5\% marked nodes to generate outliers (\textit{i.e.,} noisy samples). From the sub-figures in Fig. \ref{5}, some key observations can be concluded: 1) the proposed RARE achieves an approximate 2.5\% accuracy gain against SOTA GraphMAE on both adversarial and original datasets, which demonstrates the superiority and effectiveness of our method;
2) GraphMAE suffers from obvious performance degradation after around 400 iterations until convergence, while RARE substantially enhances the training stability of masked graph autoencoder with performance continually reaching a plateau. We attribute this to the integration of implicit and explicit self-supervision mechanisms, which can regularize each other to provide more reliable guidance for model training and produce better performance than only minimizing a raw data reconstruction loss; 
3) when GraphMAE processes a noisy graph, the average data loss value of outliers (\textit{i.e.,} the red curve) is generally smaller than that of normal nodes (\textit{i.e.,} the green curve), indicating that GraphMAE fits outliers better than normal nodes. This phenomenon is opposite to that of our method, implying that RARE has stronger robustness against adversarial attacks than GraphMAE. These results solidly support our claim that the robustness of the model would be compromised when lacking of reliable self-supervision guidance for model learning; and 4) we also notice an interesting phenomenon that the latent loss curves of outliers and normal nodes almost overlap each other on adversarial WikiCS. We guess that this is because our implicit self-supervision mechanism is driven by the MSE loss, which could carefully collect and preserve the informative features but is not selective enough to focus RARE on those outliers that are hard to tell. We will investigate this phenomenon thoroughly in our further work.

\begin{figure}[!t]
\centering
\includegraphics[width=3.6in]{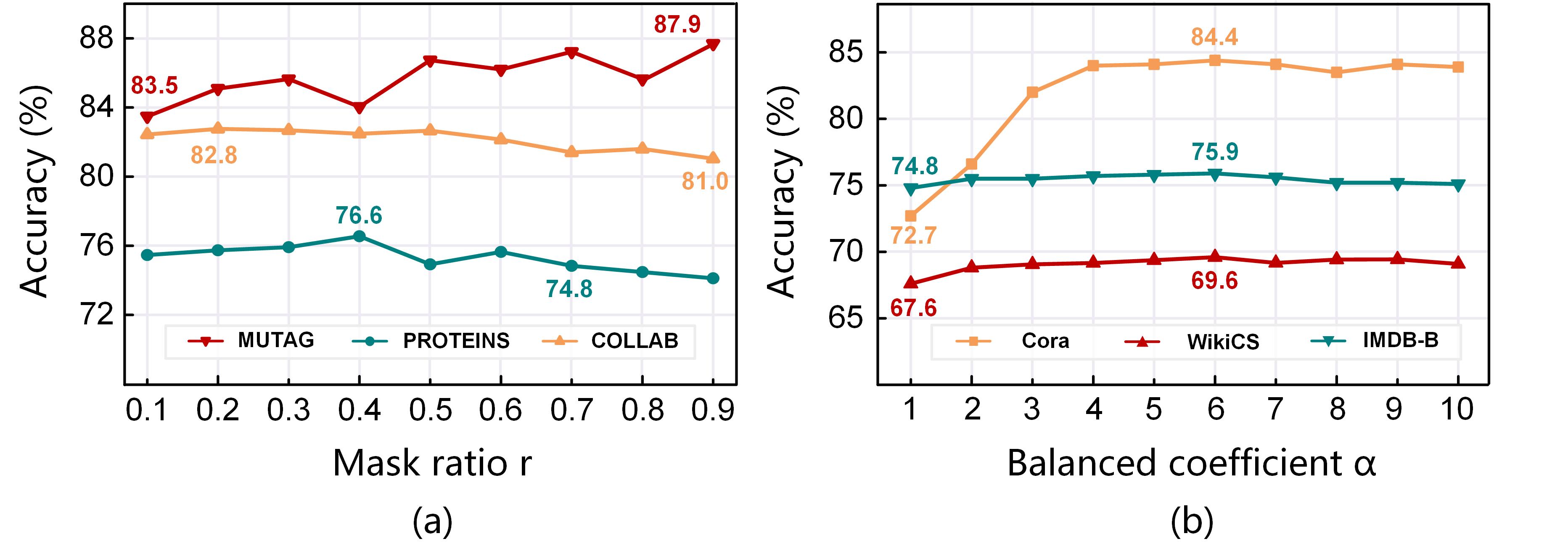}
\caption{Performance comparison with the variation of hyper-parameters: (a) The sensitivity of RARE when $r$ varies from 0.1 to 1 with 0.1 step size; (b) The sensitivity of RARE when $\alpha$ varies from 1 to 10 with 1 step size. The X-axis and Y-axis refer to the $r$ (or $\alpha$) value and the ACC performance, respectively.}
\label{6}
\end{figure}

\begin{figure}[!t]
\centering
\includegraphics[width=3.5in]{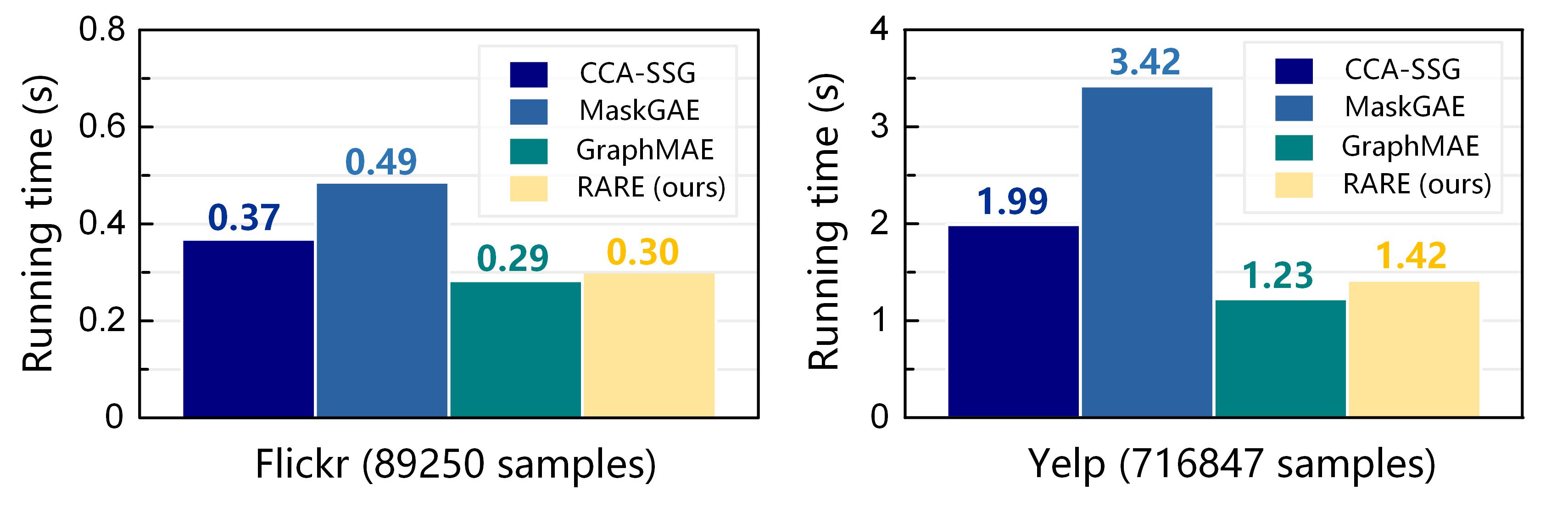}
\caption{Running time consumption (second) on Flickr and Yelp. All methods are evaluated on the same device with one NVIDIA 3090 GPU card, and the reported result refers to an average time
of 10 iterations.}
\label{7}
\end{figure}

\begin{figure}[!t]
\centering
\includegraphics[width=3.40in]{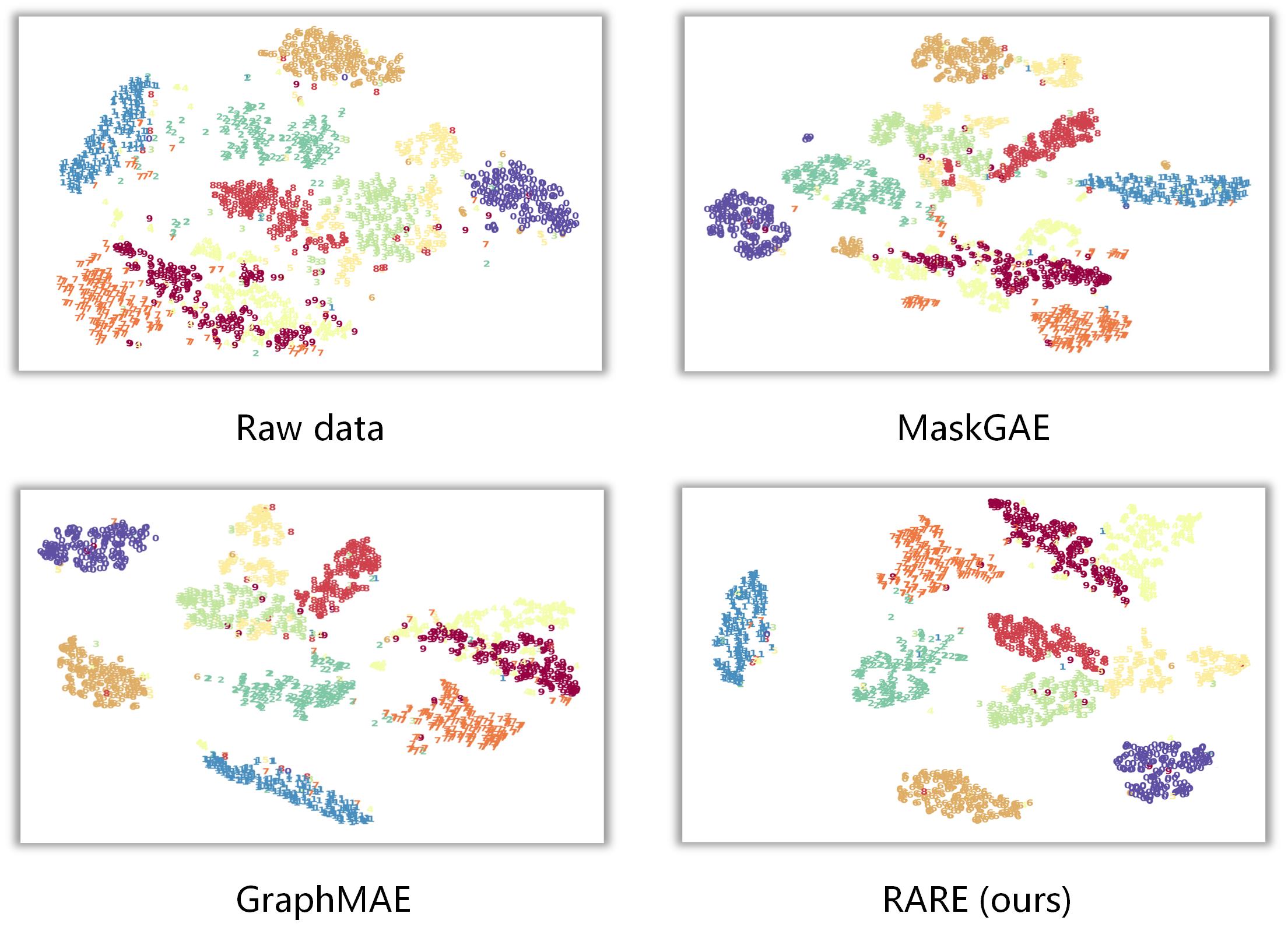}
\caption{Visualization of raw data and three MGAEs on Mnist. In our settings, we randomly sample 2,000 samples and utilize the T-SNE toolkit to visualize their raw data and latent features learned by three MGAEs, respectively. The proposed RARE presents clearer partitions and denser cluster structures than MaskGAE \cite{2022MaskGAE} and GraphMAE \cite{2022GraphMAE}.}
\label{8}
\end{figure}

\begin{figure*}[!t]
\centering
\includegraphics[width=6.75in]{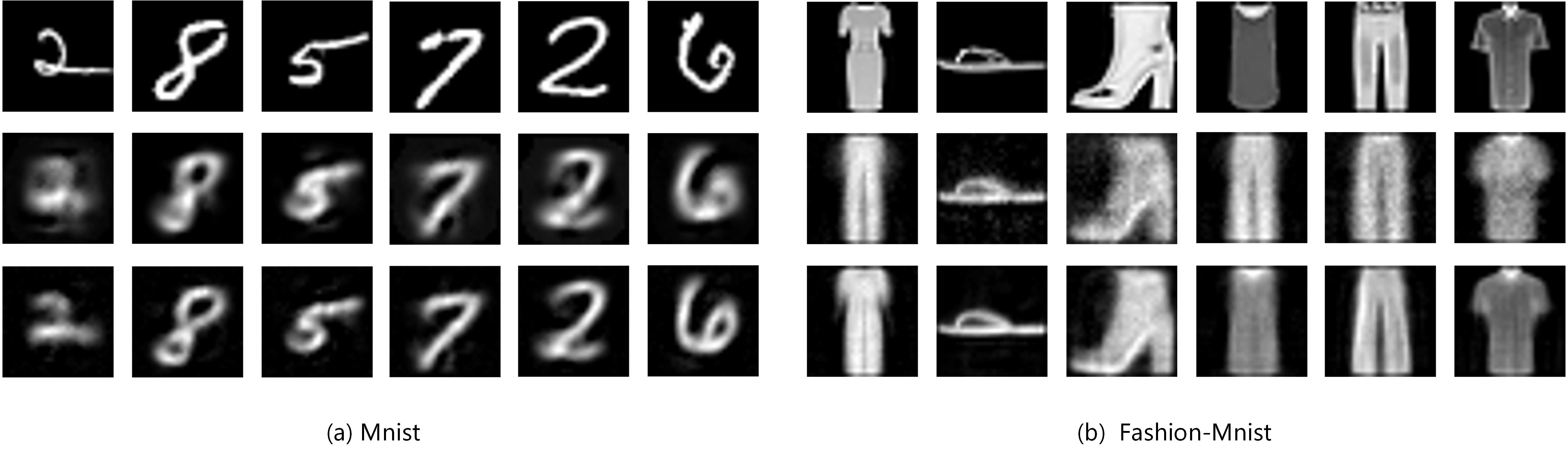}
\caption{Reconstructed image illustration: (a) Mnist; (b) Fashion-Mnist. In our settings, we first formulate the image data as a graph by regarding each image as a node and generating a KNN-based adjacency matrix, and then recover the masked nodes (\textit{i.e.,} a portion of images are entirely masked) from observations. To intuitively understand the superiority of our method, we utilize the Pillow toolkit to visualize the masked nodes (\textit{i.e.,} image samples) decoded from the latent space. The first, second, and last rows correspond to the raw data of masked images, masked images reconstructed by GraphMAE \cite{2022GraphMAE} and RARE, respectively.}
\label{9}
\end{figure*}

\subsection{Hyper-parameter Sensitivity (RQ4)}
\subsubsection{Impact of Mask Ratio $r$}
To further illustrate the superiority of RARE, we investigate its performance variation with respect to different mask ratios. Concretely, we pre-train RARE by varying the mask ratio $r$ from 0.1 to 0.9 with a step size of 0.1. From the results shown in Fig. \ref{6}(a), some key observations can be obtained: 1) taking the results on PROTEINS for example, continually increasing the value of $r$ first improves the model accuracy and then leads to relatively poor performance. This indicates that the masking-then-reconstructing mechanism is indeed effective for self-supervised graph pre-training, but a proper $r$ is required to balance the visible and masked information; 2) increasing $r$ by more than 0.6 would cause a performance drop in most cases, while the proposed RARE can still perform well within a wide range of high mask ratios. For example, the optimal masked ratio (\textit{i.e.,} 90\%) for MUTAG is surprisingly high, indicating that in some cases, the model with a high mask ratio largely eliminates redundancy and thus yields a nontrivial and meaningful self-supervision task; and 3) the performance of RARE is relatively stable across a wide range of $r$ on COLLAB. This once again implies that RARE can also learn useful representations with limited observed information, indicating its potential to achieve a good accuracy-efficiency trade-off.

\subsubsection{Impact of Hyper-parameters $\alpha$}
Eq. (\ref{eq:5}) introduces a hyper-parameter to balance the importance of two loss functions. To show its influence in-depth, Fig. \ref{6}(b) presents the accuracy variation of RARE on three datasets when $\alpha$ varies from 1 to 10 with a step size of 1. Our observations from this sub-figure are as follows: 1) the effect of tuning $\alpha$ on model performance varies across different datasets, but the stability of the model performance is higher in the range of [5, 10]. This indicates that searching $\alpha$ from a reasonable hyper-parameter region could positively influence the model performance; 2) the accuracy variation is relatively stable across a wide range of $\alpha$ on IMDB-B and WikiCS, while on Cora, it shows a trend of first rising and then dropping slightly. This suggests that RARE requires a suitable $\alpha$ to ensure the quality of learned representations when reconstructing the raw attributes; and 3) RARE tends to perform well by setting $\alpha = 6$ according to the results on all datasets. 

\subsection{Running Time Consumption (RQ5)}
Fig. \ref{7} shows a comparison of the running time consumption among a scalable contrastive SGP method (\textit{i.e.,} CCA-SSG) and three MGAEs on Flickr and Yelp datasets. All methods are evaluated on the same device with one NVIDIA-3090 GPU card, and the reported result refers to an average time of 10 iterations. 
As evidenced by the results, since RARE outperforms CCA-SSG in terms of speed on both Flickr and Yelp datasets, the computational efficiency of utilizing a masked graph autoencoder to handle large-scale graphs is promising.
Moreover, RARE can achieve better accuracy than MaskGAE and GraphMAE without considerably increasing the computation cost. These experimental results are consistent with previous complexity analyses, demonstrating that our method can scale to large-scale scenarios with linear scalability.

\subsection{Visualization (RQ6)}
In this subsection, we present comparisons of T-SNE and reconstructed images among several methods, respectively. In Fig. \ref{8}, the samples with different colors indicate different categories predicted by methods. As seen, RARE presents clearer partitions and denser cluster structures than MaskGAE and GraphMAE, especially for Category 3 and Category 5. This verifies that RARE can learn more compact and discriminative node representations. This is attributed to integrating implicit and explicit self-supervision mechanisms for masked content recovery by performing a joint mask-then-reconstruct strategy in both latent feature and raw data spaces. To intuitively understand the superiority of our method, we utilize the Pillow toolkit to visualize the raw data of masked images, the masked images reconstructed by GraphMAEA and RARE, respectively, as shown in Fig. \ref{9}. These results once again illustrate that our method can achieve a better reconstruction in the raw data space with clearer boundaries and semantics than baselines. This verifies that RARE can indeed provide a more comprehensive understanding of the original data. 

\section{Conclusion and Future Work}
In this work, we revisit the inherent distinction between traditional data formats (\textit{e.g.,} images and texts) and graphs for masked signal modeling, and investigate the applicability problem of leveraging masked autoencoders to process graph data. This motivates us to propose a novel framework called RARE for self-supervised graph pre-training. In our method, we implement both implicit and explicit self-supervision mechanisms for masked content recovery by performing a joint mask-then-reconstruct strategy in both latent feature and raw data spaces. Particularly, the designed masked latent feature completion scheme can improve the certainty in inferring masked data and the reliability of the self-supervision mechanism. We also provide theoretical analyses to explain why RARE can work well and how the designed components contribute to the model performance. Extensive experiments on seventeen datasets have demonstrated the effectiveness and superiority of RARE on three downstream tasks. 

However, there are still some limitations of existing masked graph autoencoders that have not been fully explored. For instance, existing MGAEs assume that all samples within a graph are available and complete, which may not always hold in practice since it is hard to collect all information from real-world graph data. Future work may extend the proposed RARE to the data-incomplete circumstance, and investigate the connections and differences between self-supervised masked graph pre-training and self-supervised incomplete graph pre-training. Moreover, in the current version, RARE only supports reconstructing each masked node from its adjacent neighbors via graph attention layers. In the future, developing a more effective and efficient MGAE to explore global features for information recovery is another interesting direction.

\bibliographystyle{IEEEtran}
\bibliography{mytnnls}

\end{document}